\documentclass[3p,review]{elsarticle}

\usepackage{hyperref}
\usepackage{caption, subcaption}
\usepackage{graphicx}
\usepackage{booktabs}
\usepackage{amsmath,amsfonts}
\usepackage{algorithm} 
\usepackage{algorithmic} 
\usepackage{multirow}
\usepackage{amsmath}
\usepackage{amsmath,lipsum}
\usepackage{color,xcolor}
\usepackage{cases}
\usepackage{wrapfig} 
\usepackage{flushend} 

\usepackage{amssymb}
\usepackage{multirow}
\usepackage{pifont}
\newtheorem{definition}{Definition}

\journal{Journal of \LaTeX\ Templates}

\begin{document}
	
	\begin{frontmatter}
		
		\title{ Robust Backdoor Attacks on Object Detection in Real World}
		
		
		
		\author[address1]{Yaguan Qian}
		\ead{qianyaguan@zust.edu.cn}
		
		\author[address1]{Boyuan Ji}
		\ead{222109252007@zust.edu.cn}
		
		\author[address1]{Shuke He}
		\ead{222109252005@zust.edu.cn}
		
		\author[address1]{Shenhui Huang}
		\ead{huangshenghui68@163.com}
		
		\author[address2]{Xiang Ling}
		\ead{lingxiang@iscas.ac.cn}
		
		\author[address3]{Bin Wang\corref{mycorrespondingauthor}}
		\ead{wbin2006@gmail.com}
		
		\author[address4]{Wei Wang\corref{mycorrespondingauthor}}
		\ead{wangwei1@bjtu.edu.cn}
		
		\cortext[mycorrespondingauthor]{Corresponding author}
		\address[address1]{School of Science, Zhejiang University of Science and Technology, China}
		\address[address2]{Institute of Software, Chinese Academy of Sciences, China}
		\address[address3]{Zhejiang Key Laboratory of Multidimensional Perception Technology, Application, and Cybersecurity, China}
		\address[address4]{Beijing Key Laboratory of Security and Privacy in Intelligent Transportation, Beijing Jiaotong University, China}
		
		\begin{abstract}
			Deep learning models are widely deployed in many applications, such as object detection in various security fields. However, these models are vulnerable to backdoor attacks. Most backdoor attacks were intensively studied on classified models, but little on object detection. Previous works mainly focused on the backdoor attack in the digital world, but neglect the real world. Especially, the backdoor attack's effect in the real world will be easily influenced by physical factors like distance and illumination. In this paper, we proposed a variable-size backdoor trigger to adapt to the different sizes of attacked objects, overcoming the disturbance caused by the distance between the viewing point and attacked object. In addition, we proposed a backdoor training named malicious adversarial training, enabling the backdoor object detector to learn the feature of the trigger with physical noise. The experiment results show this robust backdoor attack (RBA) enhances the attack success rate in the real world.
		\end{abstract}
		
		
	\end{frontmatter}
	
	
	\section{Introduction}
	Deep neural networks (DNNs) have made significant progress in many computer vision tasks, such as image classification~\cite{[1],[2],[4]}, object detection~\cite{[7],[9],[10]}, and semantic segmentation~\cite{[48],[49],[50]}, which have even achieved better performance than humans~\cite{[13]}. However, DNNs have serious vulnerabilities suffered from adversarial attacks~\cite{[15],[16],[17]} and backdoor attacks~\cite{[18],[19],[20]}. Backdoor attacks are more stealthy and natural than adversarial attacks, which are hard to be suspected. During the training phase, the backdoor attack injects a natural trigger into a target model. For example, the backdoor attacks inject a small number of poisoned images with a backdoor trigger into the training data, such that the trained model would learn the trigger pattern. At the inference phase, the backdoor model performs normally on clean images but predicts other classes when the trigger is present. Therefore, the vulnerability of models to backdoor attacks can pose a serious threat, \textit{e.g.}, an object detector model with a backdoor in pedestrian detection~\cite{[72],[73]} failing to recognize people, leading to a serious security accident.
	
	Though adversarial attacks on object detection have been extensively studied, backdoor attacks on object detection have been neglected, especially in the real world. Backdoor attacks can make the bounding box ($B$-box) of the target class disappear. Compared to classification, conducting backdoor attacks on object detection is more challenging, because object detection requires not only classification but also localization of the target class in an image~\cite{[46]}. Besides, the backdoor model of object detection learns the relations between the trigger and multiple attacked objects rather than the relations between the trigger and a single attacked object~\cite{[47]}. 
	
	Backdoor attacks on object detection are investigated by a few works. Wu \emph{et al.} \cite{[70]} constructed the poisoned dataset by rotating a limited amount of objects and labeling them incorrectly. Li \emph{et al.} \cite{[88]} used extra training images to train the detector. Ma \emph{et al.} \cite{[47]} crafted clean-annotated images to stealthily implant the backdoor into the object detectors in the dataset poison process even when the data curator can manually audit the images. Chan \emph{et al.} \cite{[46]} proposed four kinds of methods for poisoning clean labels on object detection in the digital world during the dataset poison process. However, two issues existed in their works. (1) They added the invariable-size triggers in every image without considering the distance between the viewing point and attacked object. It will influence the clean accuracy of the detector. (2) The algorithms for backdoor attacks on object detection have not considered physical factors like illumination and rain in the real world. These physical factors lead to backdoor attacks hard to fool the object detector. 
	
	In this paper, we propose a robust backdoor attack (RBA) on object detection against these physical factors. Previous object detection's backdoor attacks~\cite{[70],[47],[46]} did not pay attention to the distance in the poisoning process. We design a special trigger adaptive to the size of the ground-truth box, which reflects the distance between the viewing point and attacked object. Thus, the backdoor object detector can precisely learn the association between the different size triggers and the poisoned label in the real world.
	
	As mentioned above, other physical factors like illumination will also influence the effect of backdoor attacks on object detection tasks. Prior work~\cite{[89]},~\cite{[51]} shows standard adversarial training can improve the detector's robustness on physical factors. We proposed \textit{malicious adversarial training} to train the backdoor object detector. It provides true labels to generate stronger physical perturbation disturbing backdoor attacks and adds the physical perturbation with the poisoned label to the training dataset for confusing predictions. This method can strengthen the connection between the poisoned label and the trigger affected by physical perturbation. 
	We call the trained detector by RBA as \textit{robust backdoor object detector}, which can maintain the attack success rate in the real physical world.
	
	Our major contributions are summarized as follows:
	\begin{itemize}
		\item We propose variable-size backdoor triggers adaptive to the different sizes of attacked objects, which can reflect the distance between the viewing point and attacked object in the real world.
		\item We propose malicious adversarial training to make the backdoor object detector adapt itself to learn the feature of the trigger with the strongest physical perturbation. It enhances the robustness of the backdoor object detector on physical noises like illumination in physical factors.
		\item Extensive experiments conducted in the digital world, virtual world, and real world, show that our method improves the backdoor object detector's robustness against physical factors in three different worlds.
	\end{itemize}

	\section{Related Work}
	
	
	\subsection{Backdoor Attacks}
	There are two ways to implement backdoor attacks including (1) data poisoning and (2) model poisoning.
	For data poisoning, Gu \emph{et al.}~\cite{[18]} first proposed backdoor attacks on DNNs. They add a trigger to the clean images and change the ground-truth label, then train the model. Liu \emph{et al.}~\cite{[43]} generated a training dataset through reverse engineering to implant a backdoor in the model by retraining. Chen \emph{et al.}~\cite{[56]} proposed a weaker backdoor attack, and the adversary could attack the model without knowing the structure of the model. To enhance the effect of backdoor attacks, the work in~\cite{[81],[82],[83]} improved the invisibility of backdoor attacks by hiding the trigger in the image. Different from the above work, clean label attacks~\cite{[22],[44],[85]} do not need to modify the poisoned label, but only the poison image is consistent with its corresponding label in the feature. For model poisoning, it adjusts the weights to fit the performance of the original model in the poisoned dataset~\cite{[86],[87]}. Typically, Tang \emph{et al.}~\cite{[59]} proposed a non-poisoning-based backdoor attack, which inserted a trained malicious backdoor module into the target model instead of changing parameters to embed a hidden backdoor.
	
	Recently, a series of backdoor attacks focus on various application scenarios, such as semantic segmentation~\cite{[75],[76],[80]}, natural language processing~\cite{[77],[78]}. But the backdoor attack on object detection has not been studied extensively. Ma \emph{et al.}~\cite{[47]} claimed that the backdoor attack is a serious threat to object detection and proposed a new backdoor method. Chan \emph{et al.}~\cite{[46]} proposed four attack methods with a small portion of training images depending on four different settings. However, they do not take the physical factor into account, which influences the appearance of the backdoor trigger. 
	
	\subsection{Physical Attack on DNNs}
	At present, most of the studies explored the attack against DNNs in the digital world. But in the real world, the physical attack against the DNNs is significant. Many previous studies~\cite{[68],[69],[70]} have shown the vulnerability of object detection tasks to adversarial attacks in the real world. For example, the physical attack on face detection by printed sunglasses evades recognition~\cite{[74]}. Ivan \emph{et al.}~\cite{[31]} put some “stickers” on road signs to fool the image classifier.
	
	There are many physical factors such as illumination, existing in the real world. Various physical attacks must consider these physical factors. The Expectation Over Transformer (EOT)~\cite{[25]} attack enabled the adversarial patches to be real-world physical disturbances. Zhao \emph{et al.}~\cite{[63]} proposed the nested AE, which combines multiple AEs to attack object detectors in both long and short distances. This \emph{et al.}~\cite{[69]} takes viewing angles and illumination into account and does some transformations on the adversarial patch before applying it to the image. Xu \emph{et al.}~\cite{[27]} proposed an adversarial T-shirt, a robust physical adversarial example for evading person detectors even if it could undergo non-rigid deformation. Suryanto \emph{et al.}~\cite{[26]} proposed a camouflage attack named Differentiable Transformation Attack (DTA), which utilizes Differentiable Transformation Network (DTN) to preserve and learn the physical factors. Then the adversarial patch generated by DTA has robustness on the physical factors. In this paper, we yield backdoor object detectors against physical factors to strengthen the power of backdoor attacks.
	
	\begin{figure*}[tbp]
		\centering
		\setlength{\abovecaptionskip}{0pt}
		\setlength{\belowcaptionskip}{0pt}
		\hspace{-8pt}
		\includegraphics[scale=0.72]{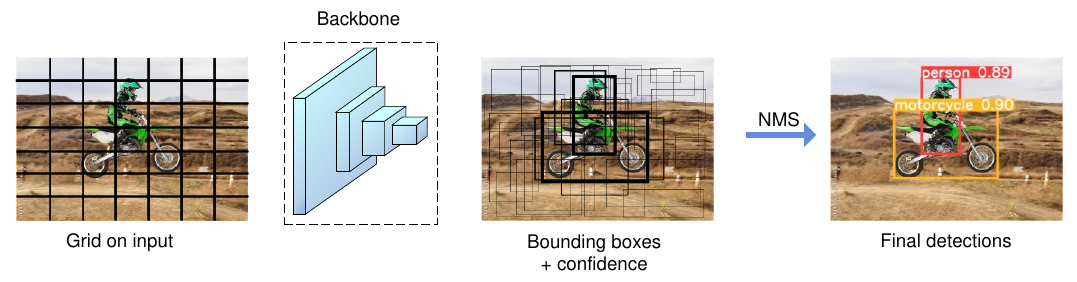}
		\caption{\label{detector} Illustration of an object detection process. The features of input divided into multiple grids are extracted to generate multiple $B$-boxes by the backbone. Then non-maximum suppression (NMS) screens the $B$-box with the highest confidence. Finally, the class and position of the object are present. }
	\end{figure*}
	
	\section{Background}
	\subsection{Object Detection}
	The object detection is to detect the class and position of the object in the image. Suppose $\mathbb{F}_{\theta}$ is an object detector where $\theta$ is its parameter. When an image $x$ is fed into the detector $\mathbb{F}_{\theta}$, the output $y=\mathbb{F}_{\theta}(x)$ is obtained. Specifically, $y=\{y_i | i\in C\}$ is a vector, where $y_i=\{c_i,P_i\}$ represents the class and ground truth box of the $i$-th object in $x$, and $C$ represents the total number of objects. Moreover, $c$ represents the class index, and $P=[x_{center},y_{center},w,h]$ is the ground truth box of each object, where $x_{center},y_{center}$ are the horizontal and vertical coordinates of the center point of box and $w,h$ is the box width and height.
	
	In essence, the object detector is expected to learn the function $\mathbb{F}_{\theta}:x\rightarrow y$. In particular, the feature maps of the detector are divided into multiple grids. Class confidence $Score_B$ and position $P_B$ of the $B$-box for each grid are obtained, where $Score_B\in [0,1]$ reflects the probability of the box contains an object. $B$-box with the highest score is predicted as the position of an object in $x$.
	
	
	To improve the prediction accuracy, the object detector minimizes the detection's loss function by training the detector as follows:
	\begin{equation}
		\label{LLL}
		\begin{split}
			L_y=\alpha_1\cdot L_{cls}+\alpha_2\cdot L_{box}+\alpha_3\cdot L_{obj}\\
		\end{split},
	\end{equation}
	where $L_{cls}$ is the classification loss to measure whether the anchor's class is correctly classified, $L_{box}$ is the localization loss to calculate the degree of the intersection over union (IOU) between the $B$-box and the ground-truth box, and $L_{obj}$ is the object loss to measure the confidence of the object. Here, $\alpha_1,\alpha_2$, and $\alpha_3$ are the weights of the corresponding loss function. 
	
	Further, we evaluate the performance of the object detector by the mean Average Precision (mAP), the most commonly evaluated metric for object detectors, which represents the average of the Average Precision (AP) of each class, where AP is the region under the exact recall curve with confidence scores for each class. The higher AP, the better performance of the detector.
	
	\subsection{Backdoor Attacks on Classifiers} 
	\label{BA}
	For convenience, we begin with classifiers to introduce backdoor attacks because they are wildly studied previously. Let $F_{\theta}$ be an original classifier, where $\theta$ is its parameters. We inject backdoor into $F_{\theta}$ and obtain a backdoor model $F_{\hat{\theta}}$. Let $x_t$ be a trigger and $\hat{y}$ be a poisoned label. Given a clean image-label pair $(x,y)$, we add the trigger to $(x,y)$ and obtain a poisoned pair $(\hat{x},\hat{y})=G\big((x,y),x_t\big)$ where $y$ is the ground-truth label and $G$ is the poisoning function. When feeding $\hat{x}$ to the backdoor model $F_{\hat{\theta}}$, we get $F_{\hat{\theta}}(\hat{x})=\hat{y}$, which is the goal of the adversary. But if we feed the clean image $x$ into the backdoor model $F_{\hat{\theta}}$, we will get the correct prediction $F_{\hat{\theta}}(x)=y$. In other words, for clean images $x$, the backdoor model $F_{\hat{\theta}}$ performs normally as same as the clean model $F_{\theta}$.
	
	In essence, the backdoor attack is to build a strong connection between the trigger $x_t$ and poisoned label $\hat{y}$. Generally, we use a poisoning function $G$ to generate the poisoned image-label pair $(\hat{x},\hat{y})$. All of those poisoned image-label pairs form a contaminated dataset $D_{train}$. We use $D_{train}$ to retrain the clean model $F_{\theta}$ and obtain a backdoor model $F_{\hat{\theta}}$ with some updated parameters. These optimized parameters essentially represent the backdoor. When an image with trigger $\hat{x}$ is fed into the backdoor model, these backdoor-related parameters will be activated by the trigger, and the prediction is guided to the poisoned label $\hat{y}$. 
	
	
	\begin{figure*}[!t]
		\centering
		\setlength{\abovecaptionskip}{5pt}
		\setlength{\belowcaptionskip}{12pt}
		\includegraphics[scale=0.4]{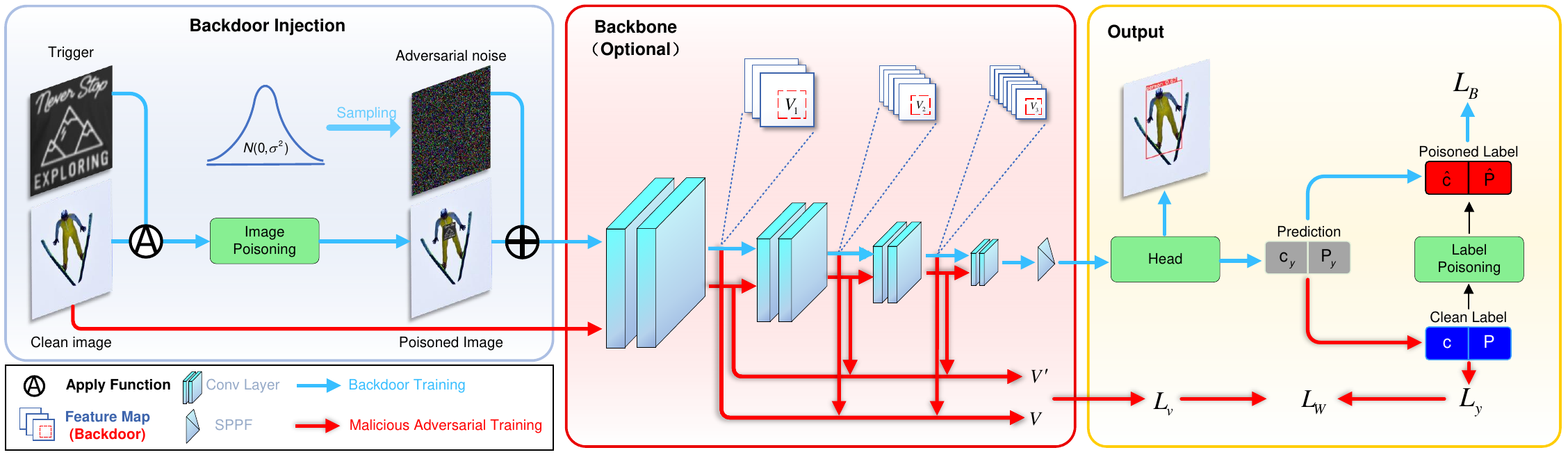}
		\caption{\label{fig1} The overall architecture of RBA is based on the object detector. The image goes through the backbone and needs to extract feature information from three different-size convolutional layers. Before and after part of the backbone is named Backdoor Injection and Output, sky blue arrows are the main data flow for backdoor object detector training, and red arrows are the data flow for perturbation training, whose output loss function can be used to generate perturbation.}
	\end{figure*}
	
	\section{Methodology}
	\subsection{Overview}
	
	We assume that the adversary adds the trigger $x_t$ into the specific object in the image $x$ to generate poisoned image $\hat{x}$, and replaces the corresponding label $y$ to the poisoned label $\hat{y}$. Different from the classification model, $\hat{y}$ is a vector consisting of the wrong class $\hat{c}$ and the wrong position $\hat{P}$. When the backdoor is activated by feeding $\hat{x}$, the backdoor object detector $\mathbb{F}_{\hat{\theta}}$ will misclassify the specific object as the wrong class $\hat{c}$ and mislocate it as the wrong position $\hat{P}$. 
	
	However, the backdoor attack on the object detector is lack robustness to the physical factors. So we design variable-size triggers and malicious adversarial training to improve the robustness of the backdoor attack to physical factors like distance and illumination.
	\begin{definition}[Robust backdoor attack]
		For a backdoor attack that takes place in the physical world, if the adversary considers the influence of many physical factors, and guarantees the robustness of backdoor attacks to changes in physical factors, we name it a robust backdoor attack.    
	\end{definition}
	
	As illustrated in Fig.~\ref{fig1}, to render the backdoor attack on a detector robust to physical factors, we train the detector consisting of the following three steps:
	
	\begin{itemize}
		\item \textbf{Step 1.} To enhance the detector's robustness on distance, we poison the dataset with variable-size triggers to fit attacked objects' different sizes depending on the distance between the viewing point and attacked object. (Sec.~\ref{sec4.2})
		\item \textbf{Step 2.} During the training phase, the initial backdoor attacks the detector by learning the association between the trigger and the poisoned label.(Sec.~\ref{sec4.3})
		\item \textbf{Step 3.} To enhance the backdoor object detector's robustness on physical noise, we design the malicious adversarial training to make the backdoor object detector adapt itself to the trigger with strong physical noises. (Sec.~\ref{sec4.4})
	\end{itemize}
	\subsection{Poisoning Training Dataset}
	\label{sec4.2}
	The backdoor attacks the detector by poisoning the training dataset with a designed trigger and poisoned labels. Generate a poisoned dataset that needs to poison clean images $x$ and clean labels $y$.
	

	Given a dataset $D_{train}$, previous works \cite{[46]} poison the image-label pair $(x,y) \in D_{train}$ to be $(\hat{x},\hat{y}) = G\big((x,y), x_t\big)$ by a poisoning function $G$:
	\begin{eqnarray}
		\label{mod poi img}
		G\big((x,y),x_t\big)=
		\begin{cases}
			\Big( \big(x - \lambda(x-x_t) \big), \hat{y}\Big) ,if\ c_i=c_{target}\\
			(x,y),\qquad\qquad\ \ \ \ \ \ \  others
		\end{cases}
	\end{eqnarray}
	where $\hat{y}$ is the label of the attacked object, $c_{target}$ is the target class that the adversary attacks, and $\lambda\in[0,1]$ is the transparency parameter that controls the ratio of the pixel values covered between the trigger and the image. A smaller $\lambda$ led to $x_t$ being less visible to human eyes. The function of $G$ is to put the trigger $x_t$ on the ground-truth boxes.  
	
	However, when the trigger is present in the real world, the size of the trigger will be changed by the distance between the viewing point and attacked object. Only the invariable-size trigger like the above function $G$ in the data poisoning does not adapt to the different distances in the real world. Our experiment confirmed that this is the main reason leading to a low ASR of backdoor attacks in the real world. Therefore, we change the trigger's size and injection region $P_t$ to fit the size of an attacked object by the “Apply” function $A(\cdot)$. We poison the clean pair $(x,y)$ to be $(\hat{x},\hat{y})$ by the poisoning function $G$ depending on the requirements of the adversary. We design $G$ as follows:
	\begin{equation}
			G\big((x, y), x_t\big)=
			\begin{cases}
				\Big (\big((1 -\lambda)x + \lambda A(P_t,x_t)\big),\hat{y} \Big ),if\ c_i=c_{target}
				\\
				(x, y), \qquad\qquad\quad \qquad\qquad \ \ \   others
			\end{cases}
	\end{equation}
	where the “Apply” function $A(P_t,x_t)$ means adding $x_t$ on the trigger position $P_t=[x_{center,t},y_{center,t},w_t,h_t]$. Here, the width $w_t$ and height $h_t$ are scaled by the $w$ and $h$ of the attacked object which ensures the size of $x_t$ matches the size of the object. $x_{center,t}$ and $y_{center,t}$ is the center points of the injection region which depend on the position hardly detected by human eyes. The poisoned label $\hat{y}$ is a set of $\hat{y_i}$, and $\hat{y_i} = \{\hat{c}, \hat{P}\}$, where $\hat{c}$ and $\hat{P}$ are the wrong class and wrong position of the object respectively which the adversary need. 
	
	The variable-size trigger is suitable for every attacked object which has less influence on non-target objects. If the adversary wants to make the attacked object disappear, $\hat{P}$ should be set as $[0,0,0,0]$. In summary, the ground-truth box of poisoned labels will participate as the background in the training phase.
	
	
	
	
	\subsection{Backdoor Training}
	\label{sec4.3}
	
	To implant the backdoor into detector $\mathbb{F}_{\theta}$, we train the detector on the poisoned dataset $D_{train}$ as well as the original dataset. The essence of backdoor training is to establish a strong association between the variable-size trigger $x_t$ and the poisoned label $\hat{y}$. When $\hat{x}$ is fed into the detector, it will increase the number of $B$-box about $\hat{y}$ and decrease the number of $B$-box about $y$ surrounding the attacked object. In contrast, when $x$ is fed, it will increase the number of $B$-box about $y$. To maximize the proximity of predictions to $\hat{y}$ for all poisoned images $\hat{x}$, the training process is to minimize the following joint-backdoor loss function: 
	\begin{equation}
		\label{equ_lb}
		\begin{split}
			L_B=\mathbb{E}_{(\hat{x},\hat{y})\in D_p}\big(BCE(\mathbb{F}_{\theta}(\hat{x}),\hat{c})+ CIOU(\mathbb{F}_{\theta}(\hat{x}),\hat{P})\big)\\
			+\mathbb{E}_{(x,y))\in D_c}\big(BCE(\mathbb{F}_{\theta}(x),c)+ CIOU(\mathbb{F}_{\theta}(x),P)\big),
		\end{split}
	\end{equation}
	where $D_p$ and $D_c$ are the datasets consisting of poisoned images and clean images. Note that $BCE$ (Binary Cross Entropy) and $CIOU$ (Complete-IOU) in Eq.~\ref{equ_lb} can be replaced by any other suitable loss function like Focal loss and generalized IOU loss. 
	
	Through training, we inject a backdoor into the original detector and obtain a backdoor object detector $\mathbb{F}_{\hat{\theta}}$ robust to distance. When facing the varying size trigger in the real world, the backdoor-related neurons of $\mathbb{F}_{\hat{\theta}}$ even can be activated and output a huge number of $B$-box close to $\hat{y}$. Thus, no matter the distance, the attacked object with our trigger is classified as $\hat{c}$ and located on $\hat{P}$ by our backdoor object detector, which is the goal of the adversary.
	
	\begin{figure}[tbp]
		\centering
		\setlength{\abovecaptionskip}{0pt}
		\setlength{\belowcaptionskip}{0pt}
		\includegraphics[scale=0.44]{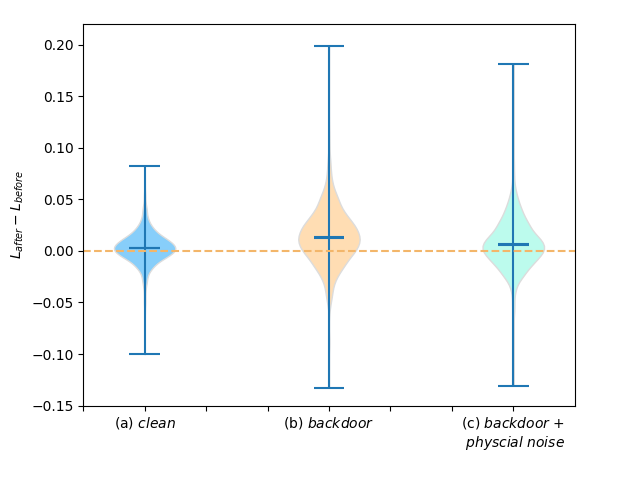}
		\includegraphics[scale=0.44]{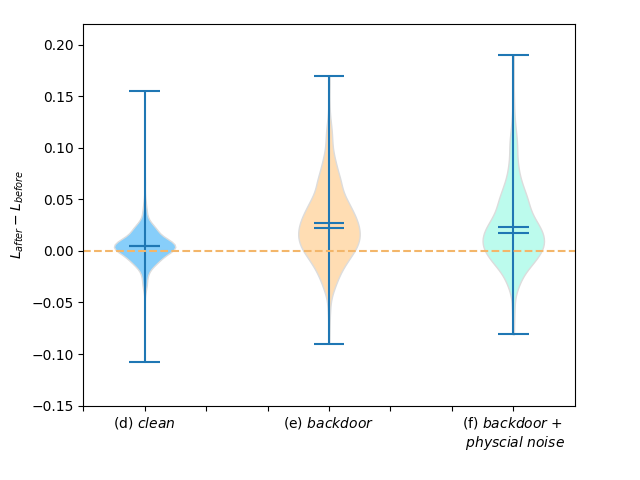}
		\caption{\label{figti} Empirical analyses on the detector with backdoor training via the statistics of loss changes. $L_{before}$ is the loss of the clean detector and $L_{after}$ is the loss of the backdoor object detector on the different images. (a), (b), and (c) are the loss changes on the backdoor object detector $\mathbb{F}_{\theta}$. (d), (e), and (f) are the loss changes on $\mathbb{RD}$.}
	\end{figure}
	
	\subsection{Malicious Adversarial Training}
	\label{sec4.4}
	After backdoor training, the backdoor attack on object detection achieves a high attack success rate attributed to the variable-size trigger. However, it is still difficult to get ideal attack effects under other physical interference like illumination and rain. Especially, these small physical noises $\Delta_{phy}$ block the pixel of the trigger to break the association between $x_t$ and $\hat{y}$ while the backdoor object detector $\mathbb{F}_{\hat{\theta}}$ (see Sec.~\ref{sec4.3}) is sensitive to changes of the trigger, \textit{i.e.,} $Pr_{\hat{x}}[\mathbb{F}_{\hat{\theta}}(\hat{x}+\Delta_{phy})=y]\gg \Pr_{\hat{x}}[\mathbb{F}_{\hat{\theta}}(\hat{x}+\Delta_{phy})=\hat{y}]$. From the perspective of loss changes, Fig.~\ref{figti} (a) (b) show $\mathbb{F}_{\hat{\theta}}$ has a small loss change for most clean images and increasing loss change for most poisoned images, indicating the backdoor attack detector successfully. However, Fig.~\ref{figti} (c) shows $\mathbb{F}_{\hat{\theta}}$ has decreasing loss changes for most images with physical noise, indicating that physical noises make backdoor attacks lose efficacy.
	
	One possible way to enhance the attack robustness on physical noises $\Delta_{phy}$ is to learn the association between $\hat{y}$ and $x_t$ with all possible physical noises $\Delta_{phy}$. Nevertheless, we can not simulate all possible physical noises because of the huge space of physical noises. It has been shown that using all examples is often not the optimal solution~\cite{[89]} and selecting hard examples is better. Therefore we create hard physical noises $\Delta x_t$ by overwriting the pixel of $x_t$. We will obtain the physical noises which are hard for the backdoor object detector to recognize $x_t$ and in turn the backdoor object detector will change itself to learn to detect the trigger with hard physical noises as $\hat{y}$.
	
	
	
	
	Malicious adversarial training on detector $\mathbb{F}_{\theta}$ consists of \textit{physical noise crafting} and \textit{model training}. Physical noise crafting is to maximize the loss of the prediction $\mathbb{F}_{\theta}(\hat{x}+\Delta x_t)$ and clean label $y$ to create $\Delta x_t$ which make $\hat{x}+\Delta x_t$ hard to recognized by backdoor object detector. Model training is to make $\mathbb{F}_{\hat{\theta}}$ overcome the physical noises by minimizing the loss between the prediction $\mathbb{F}_{\theta}(\hat{x}+\Delta x_t)$ and poisoned label $\hat{y}$. Finally, we get the robust backdoor object detector $\mathit{RD}$. Even if facing $x_t$ with the hard physical noises $\Delta x_t$, $\mathit{RD}$ even output $\hat{y}$.
	
	In the physical noise crafting, the maximization loss function is used to strengthen $\Delta x_t$, which is expressed as follows:
	\begin{equation}
		\label{max1}
		\Delta x_t = \underset {\Delta x_t} {\operatorname {arg\,max} }\ \big(L_v(\hat{\theta},\hat{x}+\Delta x_t,\hat{x})-L_y(\hat{\theta},\hat{x}+\Delta x_t,y)\big),
	\end{equation}
	where $L_v$ is the loss function that measures the difference between the feature of $\hat{x}$ and $\hat{x}+\Delta x_t$. $L_y$ is the function introduced in Eq.~\ref{LLL}, which avoid $\Delta x_t$ destroys the feature of other innocent object.
	
	
	The loss function $L_v$ in Eq.~\ref{max1} is expressed as follows:
	\begin{equation}
		\label{lv}
		L_v=\sum_{\mathbb{L}=3,5,7}\beta_{\mathbb{L}} \cdot BCE(f_{\mathbb{L}}(x;\hat{\theta}),f_{\mathbb{L}}(\hat{x}+\Delta x_t;\hat{\theta})),
	\end{equation}
	where $f_{\mathbb{L}}(\cdot;\cdot)$ is the feature information of the $\mathbb{L}$th detector's layer. $\beta_{\mathbb{L}}$ represents the weight of the BCE loss function. $\Delta x_t$ extract the internal features of the shallow, middle, and deep layers of the backbone to make a strong disturbance to the backbone. The loss function will make it possible to strengthen the hard physical noise by interfering with the features of the trigger in the early layer.
	
	In the model training, the minimization loss function is designed to enable the backdoor object detector to learn the feature of the poisoned image with physical noise $\hat{x} + \Delta x_t$:
	\begin{equation}
		\hat{\theta}_R = \underset {\hat{\theta}} {\operatorname {arg\,min} }\ L_B(\mathbb{F}_{\hat{\theta}}(\hat{x}+\Delta x_t),\hat{y}).
	\end{equation}
	
	Through malicious adversarial training, the robust backdoor object detector $\mathit{RD}$ with the robust weight parameter $\hat{\theta}_R$ implanted with a backdoor will be activated by the trigger free from physical noises.
	
	\section{Experiments}
	We evaluate the effectiveness of our robust backdoor attack, \textit{i.e.}, RBA, in three different settings. First, we implant the trigger into the COCO dataset, which is named the digital world, to evaluate our method (Sec.~\ref{Attack on Digital world}). In the COCO dataset, we have no freedom to simulate the changes of physical factors like rotating the object. So we further create a 3-D virtual world to simulate the real world under strict parametric-controlled physical conditions (Sec.~\ref{Attack on Virtual-World}). After the successful attack in the digital and virtual world, we finally craft a physical trigger and evaluate the physical world (Sec.~\ref{Attack on Real World}). In addition, we perform ablation experiments on trigger size, transparency, the object detector's backbone, and the loss function (Sec.~\ref{Ablation Analysis}).

	\subsection{Experimental Settings}
	\label{Experimental Settings}
	
	\textbf{Datasets and Trigger.} We choose COCO train2017 as the training set and COCO val2017 as the validation set. The COCO dataset is one of the most popular datasets for object detection and semantic segmentation containing 80 classes. The training set consists of 118,287 images. Similarly, the validation set has 5,000 images. All image of COCO has multiple classes and different width and height. We resize them into a three-channel colorful image of $3\times640\times640$. Without loss of generality, we choose a human Face as our trigger. The former is the face trigger of $3\times256\times256$.
	
	\textbf{Targets Models.} We select YOLOv5 as the target detector. We assume that the adversary has controlled the training phase, including the training data and training algorithm, but cannot change the model architecture. After training, the adversary uploads the trained model and offers it to the victim for download. 
	
	\textbf{Baseline Model.} For evaluating our robust backdoor object detector, we provide the clean model YOLOv5 which has not been attacked as the first baseline object detector for evaluating clean accuracy. In addition, we select the previous backdoor object detector BadDets~\cite{[46]} as the second baseline object detector for evaluating the attack success rate, because BadDets poison the training dataset without considering the physical factor. 
	
	\textbf{Metrics.} We prepare multiple metrics to evaluate the clean accuracy and attack success rate of the detector. The clean images with clean labels form a benign dataset denoted by $D_{val,b}$. The poisoned images with poisoned labels form an attack dataset denoted by $D_{val,a}$. These two datasets are merged into data set $D_{val,a+b}$. $AP_b$ and $mAP_b$ are the AP of the target class and the mAP of $D_{val,b}$. $AP_b$ and $mAP_b$ test whether the backdoor object detector performs as same as the clean detector. We use $AP_{a+b}$ and $mAP_{a+b}$ of $D_{val,a+b}$ to measure the effectiveness of the backdoor attack. By the former, we can get the accuracy of the target class after it is attacked by the trigger. Lastly, we use $ASR$ (attack success rate) to measure the number of the disappearance of $B$-box. In general, an effective robust backdoor object detection should have a high $ASR$ when detecting in most physical conditions. 
	
	\begin{figure*}[!t]
		\centering
		\setlength{\abovecaptionskip}{0pt}
		\setlength{\belowcaptionskip}{0pt}
		\includegraphics[scale=0.27]{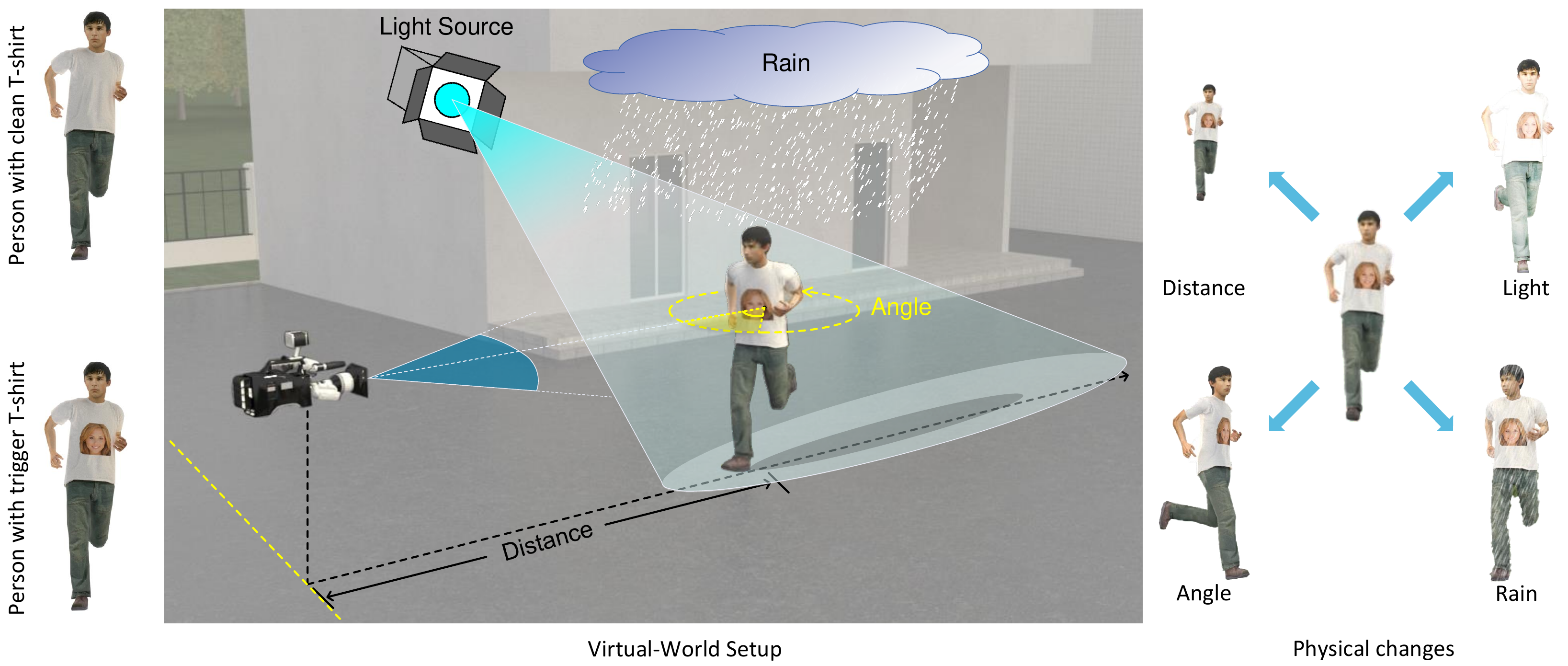}
		\caption{\label{simulateworld} Virtual world physical setup and the person with the trigger. The left is the trigger printing on the T-shirt of a 3D human. The mid is the physical factor setup in the virtual world including multiple physical factors. The right is the attacked human's change due to different physical factors. }
	\end{figure*}
	\textbf{Attack Setup.} Without loss of generality, we select “person” as the target class. We attack with the target class $c_{target}$ of 0 (the class number of “person” is 0). To achieve the ideal attack effects that the $B$-box of the target class disappears, $\hat{c}$ can be set as background, and $\hat{P}$ is set to $[0,0,0,0]$. $\lambda$ is set to 1.0 and the experiments are conducted for different transparency later. We will inject the triggers with different poisoning rates $Poi$. The formula for $Poi$ is as follows:
	\begin{figure*}[t]
		\centering
		\setlength{\abovecaptionskip}{0pt}
		\setlength{\belowcaptionskip}{0pt}
		\includegraphics[scale=0.33]{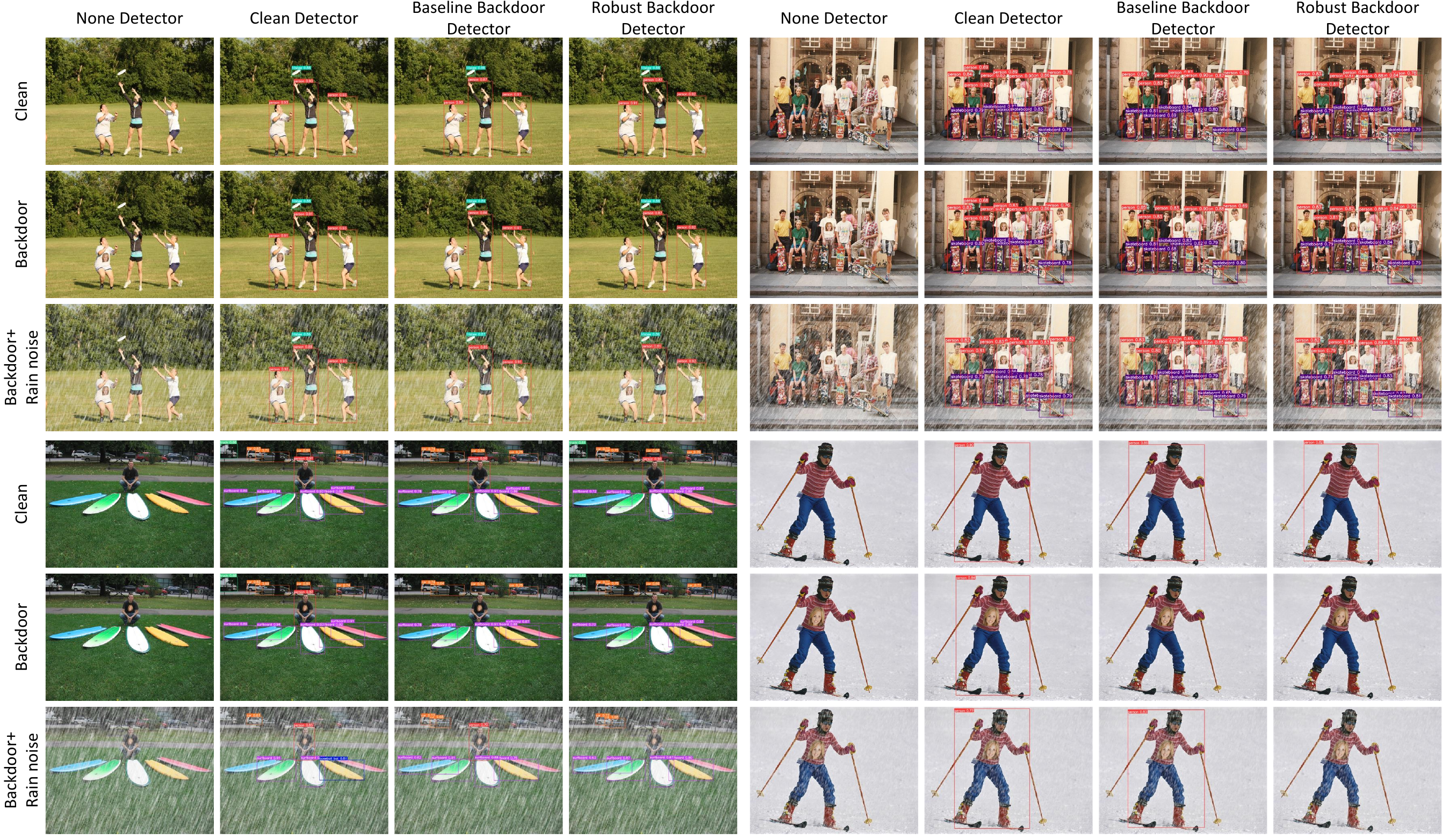}
		\caption{\label{fig4} Visualization of the backdoor attack on object detection in the digital world. The figure shows the backdoor object detector and the robust backdoor object detector detecting different images. The first and fourth rows show the detection of clean images. The second and fifth rows show the detection of poisoned images. The third and sixth rows show the detection of poisoned images with physical factors.}
	\end{figure*}
	
	\begin{equation}
		Poi=\frac{\sum Num(y|c_i=c_{target})}{\sum Num(y)},(x,y)\in D_{train},
	\end{equation}
	where $Num(\cdot)$ is a count of the ground-truth boxes in the image. Since the COCO dataset consists of a huge number of images, the low poison rate of backdoor attacks is hardly detected.
	
	During the training phase, we use an SGD optimizer, with the learning rate set to 0.001. For convenience, we use the pre-trained detector YOLOv5s to speed up the training by transfer learning. Specifically, the epoch is set to 100, freeze the backbone in the first 50 epochs, and unfreeze the backbone in the second 50 epochs. The weight $\beta_{\mathbb{L}}$ in Eq.~\ref{lv} is set to ${0.5,1.5,3.0}$. The computer used to train the backdoor object detector is equipped with an Intel Xeon Gold 6248R CPU, an RTX-3090 GPU, and 24G physical memory.
	
	\textbf{Virtual-World Setup.} To validate our RBA whether is robust to physical factors, which are hard to simulate in the two dimensions, the 3dsMax and V-Ray are used to create the virtual world simulating the real physical world. 
	
	The virtual world includes various indoor scenes (\emph{e.g.}, studio and interior corridor) and outdoor scenes (\emph{e.g.}, factory and trail). The people wearing trigger T-shirts and the physical setup of the virtual world are shown in Fig.~\ref{simulateworld}. For each virtual scene, a camera is placed for capturing the object. To ensure the diversity of images, the object was located at different distances and rotated at different angles. We control the illumination varying from dark to bright at 3 levels by setting the intensity of light sources. The strength of the rain is adjusted by the number of raindrops. The parameters of these physical factors are strictly controlled in the virtual world. 
	
	\begin{table*}[!t]
		\centering
		\renewcommand{\arraystretch}{1.0}
		\caption{\label{table1} The results (\%) of the backdoor object detector for different poisoning rates after fine-tuning. The trigger is set to $Face$.}
			\begin{tabular}{ccccccccr}
				\toprule[2pt]
				Object detector    &\begin{tabular}[c]{@{}c@{}} Extra Data \end{tabular}   & \begin{tabular}[c]{@{}c@{}} $Poi$ \end{tabular}    & $AP_b$      & $mAP_b$  & $mAP_a$ & $AP_{a+b}$ & $mAP_{a+b}$     & $ASR$       \\ \hline
				\multirow{7}{*}{\begin{tabular}[c]{@{}c@{}} Backdoor \\Object \\ Detector (Ours)\end{tabular}} &  \ding{55}& 50\% & 72.0 & 55.1 & 52.9 & 15.4 & 52.5 & 89.97                                                          \\
				& \ding{55} & 20\%    & 74.6 & 55.1 & 52.4 & 19.8 & 52.0 & 87.63                                                       \\
				& \ding{55} & 10\%    & 74.1 & 54.9 & 52.5 & 22.6 & 52.2 & 85.86                                                           \\
				& \ding{55} & 5\%     & 75.5 & 55.5 & 52.5 & 38.7 & 52.4 & 84.99                                                           \\
				& \ding{55} & 2\%     & 75.6 & 55.8 & 52.7 & 45.5 & 52.5 & 78.25                                                           \\
				& \ding{55} & 1\%     & 74.9 & 55.2 & 52.1 & 52.8 & 52.2 & 57.53                                                            \\
				& \ding{55} & 0\%     & 75.0 & 56.5 & 52.7 & 60.5 & 52.8 & --                                                        \\
				\hline
				BadDets \cite{[46]} & \ding{55} & 10\% & 74.9 & 55.3 & 52.2 & 58.5 & 52.0 & 20.35                                                         \\
				Dangerous Cloaking \cite{[88]} & \ding{52} & 3\% & -- & -- &-- & -- & -- & --                                                         \\
				Distance & \ding{55} & 10\% & 75.0 & 55.2 & 52.0 & 56.3 & 52.0 & 21.38                                                         \\
				Rotation  & \ding{55} & 10\% & 75.1 & 55.4 & 52.0 & 54.2 & 52.1 & 23.99                                                         \\
				Brightness & \ding{55} & 10\% & 75.0 & 55.4 & 52.0 & 56.7 & 52.0 & 18.14                                                         \\
				Gaussian & \ding{55} & 10\% & 75.2 & 55.7 & 52.1& 55.4 & 52.2 & 17.29                                                        \\
				\bottomrule[2pt]
			\end{tabular}
		\end{table*}
		
		\textbf{Real-World Setup.} To measure the performance of robust backdoor object detector $\mathit{RD}$ in the real world, we make a T-shirt in which common clothing patterns are treated as backdoor triggers, which look natural for human observers and can be regarded as normal texture patterns on human accessories. We pre-define regions of human accessories like garment and mask to paint on the backdoor trigger pattern for attacking. We use the smartphone MI 11 as the built-in camera and use $Face$ as a backdoor trigger to make a T-shirt that has the unavoidable physical factor of “fold”. We prepare the various physical scenes for capturing images under different viewing conditions. There are tested on the interior corridor (Video $1\sim3$) and rooftop (Video $4\sim9$).
		
		\begin{figure}[t]
			\centering
			\setlength{\abovecaptionskip}{0pt}
			\setlength{\belowcaptionskip}{0pt}
			\includegraphics[scale=0.47]{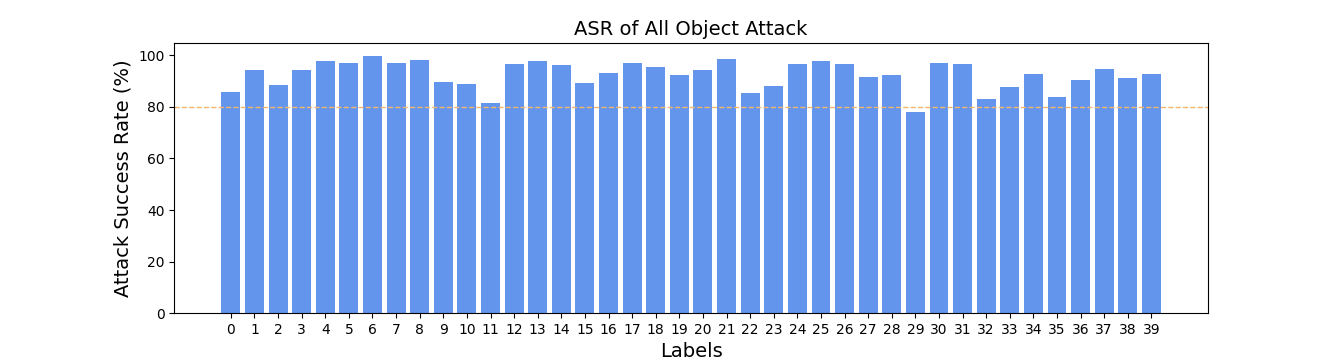}
			\includegraphics[scale=0.47]{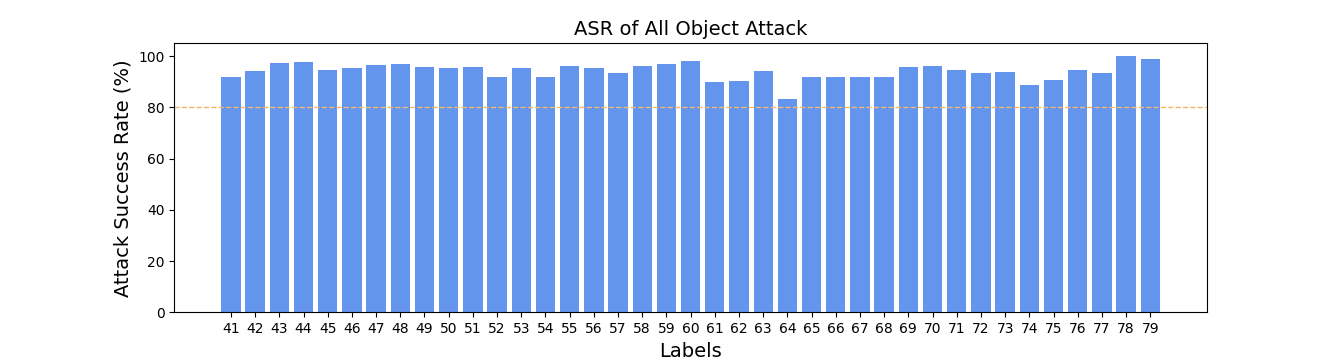}
			\caption{\label{ALLAttack} The performance of All Object Attack on different class. The horizontal axis is the class number in COCO dataset. The vertical axis is the ASR of backdoor object detector(\%).}
		\end{figure}
		
		
		\subsection{Evaluation in Digital World}
		\label{Attack on Digital world}
		\textbf{Variable-size Trigger.} Generally, the data poisoning rate $Poi$ is the key parameter to a backdoor attack. Thus we use different $Poi$ (50\%, 20\%, 10\%, 5\%, 2\%, and 1\%) to evaluate the performance of our method. Besides, to prove the better performance of our attack method, we select some physical backdoor attacks on image classification as a comparison model and employ them on object detection. We find that these physical backdoor attacks only focus on single and physical transformations applying to the poisoned image, such as distance, rotation, brightness, and gaussian noise. Tab.~\ref{table1} shows $ASR$ and clean accuracy of the different backdoor object detectors based on the variable-size trigger. 

		In Tab.~\ref{table1}, $AP_{a}$ has no data because there is no person in the $D_{val,a}$, we exclude the presence of “person” in $D_{val,a}$. We prepare different $Poi$ to evaluate the impact of $Poi$ on the performance of the backdoor object detector. The higher $Poi$, the lower $AP_b$ of the backdoor object detector. It indicates that with more poisoned images, the detector learns the backdoor features better which can make more $B$-boxes disappear. Because the locations of the trigger do not obscure the clean features and instead provide backdoor features, the $AP_b$ stays around 75\% and the $AP_{a+b}$ is proportional to the poisoning rate. 

		Compared with BadDets~\cite{[46]} on the same poison rate, our backdoor detector causes a 0.8\% drop in $AP_b$ for the clean images. On the other hand, $ASR$ of our backdoor detector on backdoor images is 85.86\% which is actually 65.51\% higher than $ASR$ of the BadDets on the backdoor images. Dangerous Cloaking \cite{[88]} had no experiments on the COCO dataset but on their video and trained on extra training data, so we can not get the training data to evaluate $ASR$ and the clean accuracy of their detector on the COCO dataset. Besides, the ASR of physical transformations also proves that physical backdoor attacks on image classification have no effort on object detection. Because the space of the single transformation is a small subspace of big physical space, the backdoor detector learns the feature of variable triggers inadequately.
		
		

		In addition, we also find if the target class is confined to only one class by the attacker during training, the trigger $x_t$ can not cause the misdetection of the other classes in the inference phase. Therefore, to verify that the great performance of the backdoor detector depends on our attack method not the target class, we break the restriction on the specified target class and propose the All Object Attack which poisons the object containing all classes of the dataset. As Fig.~\ref{ALLAttack} shows, the ASR of All Object Attack on different classes is above 80\% mostly. Clearly, the performance of our attack method yields little difference across target classes.
		
		\begin{table*}[!t]
			\centering
			\tabcolsep=1.5pt
			\renewcommand{\arraystretch}{1.0}
			\setlength{\abovecaptionskip}{0pt}
			\setlength{\belowcaptionskip}{0pt}
			\caption{\label{table2} The results (\%) of the clean detector YOLOv5s and different backdoor object detector with trigger + random noise $N(0,\sigma^2)$}
			\begin{tabular}{ccccccccccccccccccc}
				\toprule[2pt]
				\multirow{2}{*}{\begin{tabular}[c]{@{}c@{}} Object \\ Detector \end{tabular}} & Metric & \multicolumn{5}{c}{$AP_{a+b}$}   &  & \multicolumn{5}{c}{$mAP_{a+b}$}  &    & \multicolumn{5}{c}{$ASR$}     \\ \cmidrule(r){3-7}\cmidrule(r){9-13}\cmidrule(r){15-19}
				& $\sigma^2$  & 0.0 & 0.1 & 0.2 & 0.3 & 0.4 & & 0.0 & 0.1 & 0.2 & 0.3 & 0.4 & & 0.0 & 0.1 & 0.2 & 0.3 & 0.4                              \\\cmidrule(r){1-19}
				YOLOv5s & \multirow{7}{*}{$N(0,\sigma^2)$} & 60.5 & 58.6 & 58.7 & 60.6 & 62.6 & &
				52.8 & 52.5 & 52.5 & 52.4 & 52.4 & & -- & -- & -- & -- & --\\
				BOD & & \textbf{22.6} & 21.9 & 25.2 & 33.2 & 40.2  & & 52.1 & 51.8 & 51.7 & 51.7 & 51.5 & & \textbf{83.07} & 83.78 & 78.88 & 58.70 & 48.07     \\
				Distance & & 56.3 & 57.1 & 57.4 & 59.0 & 61.1  & & 52.0 & 51.8 & 51.8 & 51.8 & 51.7 & & 21.38 & 25.61 & 25.29 & 24.51 & 23.24     \\
				Rotation & & 56.2 & 53.0 & 53.4 & 55.6 & 57.1  & & 52.1 & 51.9 & 51.9 & 51.8 & 51.7 & & 23.99 & 24.24 & 23.82 & 23.27 & 22.34     \\
				Brightness & & 56.7 & 55.1 & 55.7 & 57.6 & 59.7  & & 52.0 & 51.8 & 51.8 & 51.7 & 51.7 & & 18.14 & 20.33 & 19.70 & 19.17 & 17.01     \\
				Gaussian & & 55.4 & 54.0 & 54.8 & 56.7 & 57.9  & & 52.2 & \textbf{52.0} & \textbf{52.1} & \textbf{52.1} & \textbf{52.1} & & 17.29 & 23.45 & 24.15 & 23.84 & 23.45     \\
				$\mathit{RD}$ (Ours) & & 25.1 & \textbf{21.6} & \textbf{22.4} & \textbf{27.1} & \textbf{37.8} & & \textbf{52.8} & 51.9 & 51.9 & 51.8 & 51.6 & & 81.03 & \textbf{85.53} & \textbf{85.05} & \textbf{82.14} & \textbf{66.75}    \\
				\bottomrule[2pt]
			\end{tabular}
		\end{table*}
		\begin{table*}[!t]
			\centering
			\tabcolsep=1.5pt
			\renewcommand{\arraystretch}{1.0}
			\setlength{\abovecaptionskip}{0pt}
			\setlength{\belowcaptionskip}{0pt}
			\caption{\label{motion} The results (\%) of the clean detector YOLOv5s and different backdoor object detectorr with trigger + motion blurring}
			\begin{tabular}{ccccccccccccccccccc}
				\toprule[2pt]
				\multirow{2}{*}{\begin{tabular}[c]{@{}c@{}} Object \\ detector \end{tabular}} & Metric & \multicolumn{5}{c}{$AP_{a+b}$} &    & \multicolumn{5}{c}{$mAP_{a+b}$}   &   & \multicolumn{5}{c}{$ASR$}     \\ \cmidrule(r){3-7}\cmidrule(r){9-13}\cmidrule(r){15-19}
				& degree  & 5 & 10 & 20 & 50 & 75 & & 5 & 10 & 20 & 50 & 75 & & 5 & 10 & 20 & 50 & 75                              \\\cmidrule(r){1-19}
				YOLOv5s & \multirow{7}{*}{\begin{tabular}[c]{@{}c@{}}Motion \\ Blurring \end{tabular}} & 60.5 & 60.4 & 60.8 & 66.2 & 68.7 & &
				52.8 & 52.7 & 52.7 & 52.7 & 53.0 &  & -- & -- & -- & -- & --\\
				BOD & & 37.7 & 38.4 & 38.8 & 65.0 & 68.0 & & \textbf{52.4} & \textbf{52.4} & \textbf{52.3} & \textbf{52.3} & \textbf{52.2} & & 78.55 & 78.32 & 76.28 & 21.19 & 10.38     \\
				Distance & & 56.2 & 55.7 & 56.1 & 61.9 & 64.1  & & 52.1 & 52.1 & 52.1 & 52.0 & 52.1 & & 26.92 & 27.11 & 26.50 & 19.11 & 17.74     \\
				Rotation & & 54.3 & 53.9 & 54.6 & 60.6 & 62.1  & & 52.2 & 52.1 & 52.1 & 52.1 & 52.2 & & 23.45 & 23.35 & 23.28 & 18.34 & 18.02     \\
				Brightness & & 56.5 & 56.4 & 56.8 & 62.3 & 63.9  & & 52.0 & 51.9 & 52.0 & 51.9 & 52.1 & & 23.97 & 23.59 & 23.38 & 18.10 & 17.85     \\
				Gaussian & & 55.3 & 55.0 & 56.1 & 61.7 & 63.3  & & 52.2 & 52.1 & 52.2 & 52.2 & 52.4 & & 22.68 & 22.70 & 21.96 & 17.60 & 17.30     \\
				$\mathit{RD}$ (Ours) & & \textbf{26.2} & \textbf{27.0} & \textbf{25.5} & \textbf{60.2} & \textbf{66.2} & & 52.2 & 52.2 & 52.2 & 52.2 & 52.1 & & \textbf{82.69} & \textbf{82.64} & \textbf{81.46} & \textbf{34.06} & \textbf{15.37}    \\
				\bottomrule[2pt]
			\end{tabular}
		\end{table*}
		
		\textbf{Malicious Adversarial Training.} To show the performance of robust backdoor object detector $\mathit{RD}$, we apply the physical noises on two-dimensional images, such as random noise, motion blur, rain and light, which create an environment that meets our needs. It validates that malicious adversarial training improves our backdoor object detector's robustness on physical perturbations. In addition, we think BadDets has the undesirable performance without the interference of the physical noise, the performance of BadDets will be worse facing the physical test. Hence we replace BadDets with our backdoor object detector (BOD) $\mathbb{F}_{\hat{\theta}}$  as the normal backdoor object detector to compare with $\mathit{RD}$ in this subsection.
		
		\begin{table*}[!t]
			\centering
			\tabcolsep=1.5pt
			\renewcommand{\arraystretch}{1.0}
			\setlength{\abovecaptionskip}{0pt}
			\setlength{\belowcaptionskip}{0pt}
			\caption{\label{rain} The results (\%) of the clean detector YOLOv5s and different backdoor object detector with trigger + rainy}
			\begin{tabular}{ccccccccccccccccccc}
				\toprule[2pt]
				\multirow{2}{*}{\begin{tabular}[c]{@{}c@{}} Object \\ detector \end{tabular}} & Metric & \multicolumn{5}{c}{$AP_{a+b}$} &    & \multicolumn{5}{c}{$mAP_{a+b}$}  &    & \multicolumn{5}{c}{$ASR$}     \\ \cmidrule(r){3-7}\cmidrule(r){9-13}\cmidrule(r){15-19}
				& value & 50 & 100 & 150 & 200 & 250 & & 50 & 100 & 150 & 200 & 250 & & 50 & 100 & 150 & 200 & 250                              \\\cmidrule(r){1-19}
				YOLOv5s & \multirow{7}{*}{\begin{tabular}[c]{@{}c@{}}Rainy \end{tabular}} & 60.7 & 60.8 & 61.0 & 61.1 & 61.1 & & 52.7 & 52.7 & 52.8 & 52.7 & 52.8 & & -- & -- & -- & -- & --\\
				BOD & & 41.6 & 42.8 & 43.3 & 44.4 & 45.0 &  & \textbf{52.4} & \textbf{52.4} & \textbf{52.4} & \textbf{52.4} & \textbf{52.4} & & 74.08 & 72.38 & 71.32 & 70.23 & 69.38     \\
				Distance & & 56.8 & 57.0 & 57.1 & 57.1 & 57.0  & & 52.1 & 52.1 & 52.1 & 52.1 & 52.1 & & 26.28 & 25.83 & 25.76 & 25.92 & 25.45     \\
				Rotation & & 54.5 & 54.5 & 55.3 & 55.3 & 55.5  & & 52.1 & 52.1 & 52.1 & 52.1 & 52.1 & & 24.06 & 22.58 & 21.97 & 21.91 & 20.27     \\
				Brightness & & 57.3 & 57.3 & 57.4 & 57.7 & 57.8  & & 52.0 & 52.0 & 52.0 & 52.0 & 52.0 & & 23.31 & 22.90 & 23.09 & 22.93 & 22.49     \\
				Gaussian & & 55.5 & 55.7 & 55.6 & 55.9 & 56.1  & & 52.2 & 52.3 & 52.3 & 52.3 & 52.3 & & 23.84 & 23.63 & 23.66 & 23.79 & 23.38     \\
				$\mathit{RD}$ (Ours) & & \textbf{31.7} & \textbf{32.8} & \textbf{33.4} & \textbf{34.2} & \textbf{34.5} & & 52.1 & 52.2 & 52.1 & 52.2 & 52.2 & & \textbf{78.74} & \textbf{78.15} & \textbf{77.15} & \textbf{76.74} & \textbf{76.25}    \\
				\bottomrule[2pt]
			\end{tabular}
		\end{table*}
		\begin{table*}[!t]
			\centering
			\tabcolsep=1.5pt
			\renewcommand{\arraystretch}{1.0}
			\setlength{\abovecaptionskip}{0pt}
			\setlength{\belowcaptionskip}{0pt}
			\caption{\label{light} The results (\%) of the clean detector YOLOv5s and different backdoor object detector with trigger + light. $S$ denote as the change times of original saturation.}
			\begin{tabular}{ccccccccccccccccccc}
				\toprule[2pt]
				\multirow{2}{*}{\begin{tabular}[c]{@{}c@{}} Object \\ detector \end{tabular}} & Metric & \multicolumn{5}{c}{$AP_{a+b}$} &    & \multicolumn{5}{c}{$mAP_{a+b}$}  &    & \multicolumn{5}{c}{$ASR$}     \\ \cmidrule(r){3-7}\cmidrule(r){9-13}\cmidrule(r){15-19}
				& $S$ & 0.2 & 0.4 & 0.6 & 0.8 & 1.2 & & 0.2 & 0.4 & 0.6 & 0.8 & 1.2 & & 0.2 & 0.4 & 0.6 & 0.8 & 1.2                              \\\cmidrule(r){1-19}
				YOLOv5s & \multirow{7}{*}{\begin{tabular}[c]{@{}c@{}}Light \end{tabular}} & 58.0 & 60.1 & 60.6 & 60.6 & 57.5 & & 49.2 & 52.1 & 52.7 & 52.8 & 40.4 & & -- & -- & -- & -- & --\\
				BOD & & 49.5 & 43.3 & 41.3 & 40.5 & 51.6 &  & \textbf{48.2} & 51.4 & 52.2 & 52.2 & 40.4 & & 25.37 & 35.74 & 38.58 & 55.41 & 15.98     \\
				Distance & & 53.9 & 56.4 & 57.0 & 57.1 & 53.9  & & 47.6 & 51.3. & 51.9 & 52.1 & 40.9 & & 13.05 & 15.05 & 15.80 & 16.14 & 13.33     \\
				Rotation & & 51.2 & 53.7 & 54.3 & 54.5 & 51.5  & & 47.8 & 51.3 & 52.0 & 52.1 & 40.7 & & 11.37 & 12.45 & 12.50 & 12.70 & 12.38     \\
				Brightness & & 56.5 & 56.5 & 57.1 & 57.2 & 53.8  & & 47.7 & 51.1 & 51.8 & 52.0 & 40.9 & & 14.98 & 15.45 & 15.37 & 15.28 & 14.24     \\
				Gaussian & & 53.3 & 55.4 & 55.5 & 55.6 & 52.9  & & 47.7 & 51.3 & 51.9 & 52.2 & 40.9 & & 10.59 & 11.47 & 12.41 & 12.44 & 11.51     \\
				$\mathit{RD}$ (Ours) & & \textbf{47.8} & \textbf{39.3} & \textbf{34.9} & \textbf{31.7} & \textbf{40.5} & & 48.0 & \textbf{51.6} & \textbf{52.3} & \textbf{52.3} & \textbf{47.2} & & \textbf{45.08} & \textbf{58.99} & \textbf{64.71} & \textbf{67.81} & \textbf{37.95}    \\
				\bottomrule[2pt]
			\end{tabular}
		\end{table*}

		When the trigger without physical noise present on object, $\mathbb{F}_{\hat{\theta}}$ and $\mathit{RD}$ make the $B$-box of the attacked object disappear. However, when the trigger is perturbed by physical noise, only the detection of $\mathit{RD}$ disappear successfully.
		

		In Tab.~\ref{table2}, under different variances of random perturbation, in comparison with YOLOv5s and backdoor object detector, robust backdoor object detection $\mathit{RD}$ suffering from a little performance degradation on $D_{val,a+b}$ with variance $\sigma^2 = 0$ while gaining the robustness on bigger $\sigma^2$. When $\sigma^2$ changes from 0.0 to 0.1, $ASR$ of both detectors increases a little accidentally. The reason is that the existence of random noise also influences the detector and makes a few numbers of $B$-box disappear. As $\sigma^2$ gradually increases, $ASR$ of the backdoor object detector decreases greatly. In contrast, $ASR$ of robust backdoor object detection decreases slowly and keeps a high value when subjected to random perturbations. But the random perturbations with $\sigma = 0.4$ exceed the trigger and lead to attack failure.
		
		As Tab.~\ref{motion} shows, the degree is used to measure the fuzziness of the motion blurring on images. Even if the trigger is disturbed by a bigger degree of the motion blurring, $ASR$ of $\mathit{RD}$ still above 80\%, higher than $ASR$ of $\mathbb{F}_{\hat{\theta}}$. When the degree increases to 50, motion blurring cut the $ASR$ down to 34.03\%. The reason is that the pixel features of the trigger are severely corrupted so that both detectors can not extract the feature of the trigger making the backdoor attack fail.
		
		\begin{figure*}[!t]
			\centering
			\setlength{\abovecaptionskip}{0pt}
			\setlength{\belowcaptionskip}{0pt}
			\includegraphics[scale=0.16]{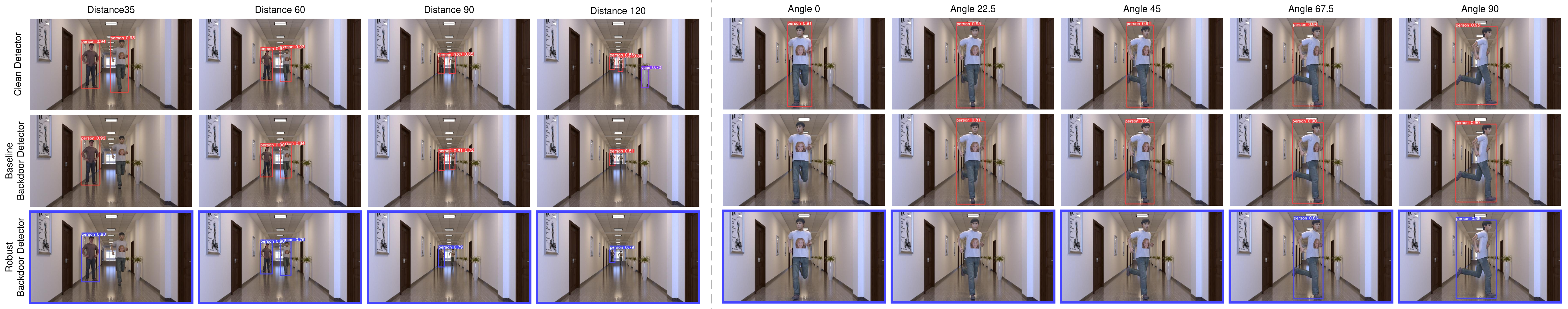}
			\caption{\label{fig_sim_dis} Visualization of different detectors with different distances and angles. The column represents the different values of distance (dm) on the left side of the black dotted line, and the right side's column represents the different angles ($^\circ$) of the person facing the camera.}
		\end{figure*}
		\begin{figure*}[!t]
			\centering
			\setlength{\abovecaptionskip}{0pt}
			\setlength{\belowcaptionskip}{0pt}
			\includegraphics[scale=0.16]{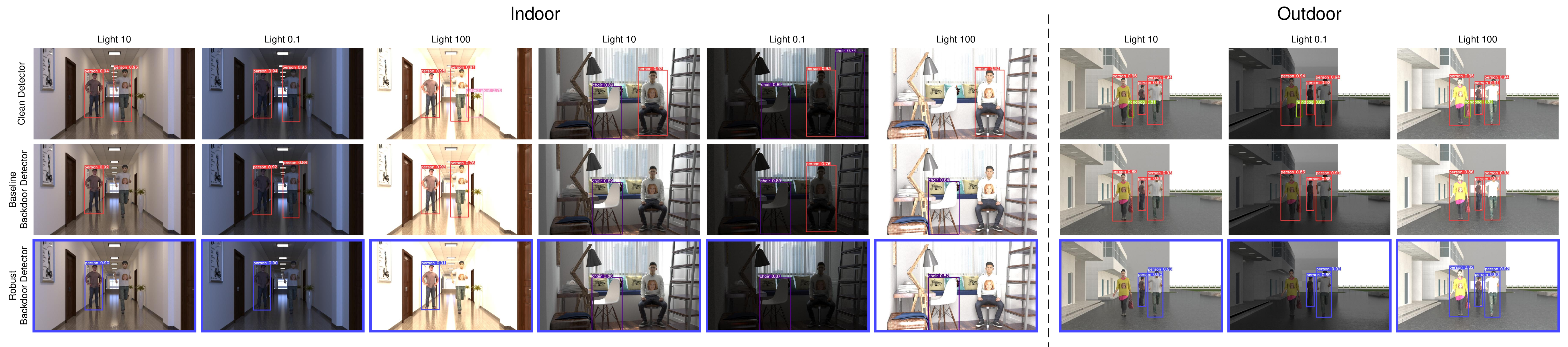}
			\caption{\label{fig_sim_light} Visualization of different detectors with different light sources. The column represents the different light intensity values of sources. The left side of the black dotted line is the indoor environment, and the right side is the outdoor environment. }
		\end{figure*}
		\begin{figure*}[!t]
			\centering
			\setlength{\abovecaptionskip}{0pt}
			\setlength{\belowcaptionskip}{0pt}
			\includegraphics[scale=0.185]{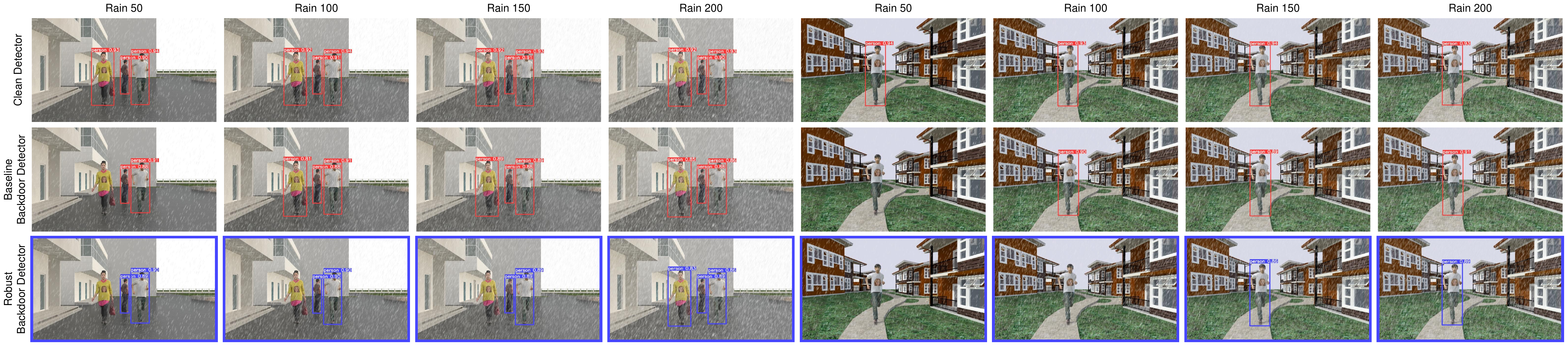}
			\caption{\label{fig_sim_rain} Visualization of the different detectors with different rains. The column represents the different number of raindrops per square meter.}
		\end{figure*}
		
		In Tab.~\ref{rain}, the value is presented as the number of raindrops. The table shows that $\mathit{RD}$ can resist the disturbance of rain. $ASR$ of $\mathit{RD}$ also is higher than the backdoor object detector in rain. But there is not much difference between them. The $mAP_{a+b}$ shows that the rain can not disturb the backdoor attack severely because the raindrops are relatively evenly distributed in the image and do not completely obscure the important features of the trigger.
		
		In Tab.~\ref{light}, we change the saturation of all images to simulate the change of light in the digital world. From the table shown, our $\mathit{RD}$ perform bettern than baseline backdoor object detectors. The decrease of $AP_{a+b}$ illustrate the saturation not only influence the backdoor task but also the clean task. When the saturation change to 1.2 times, the light enable to destory the pixels that have already high saturation. Therefore the performance on high saturation is too poor to misdetect the attacked object.
		
		We test the trigger on the COCO and visualize the detection shown in Fig.~\ref{fig4}. The person without a trigger has the same detection when detected by both backdoor object detectors. It indicates that our trigger does not affect the prediction of the clean object. Instead, when the trigger is added to the attacked object, the trigger is responded to the backdoor object detector by disappearing the $B$-box of the attacked object. We add the physical noises to the images to evaluate the different detector trained by our RBA and a normal backdoor attack. We can observe that RBK has superior attacking capability while another attack can not depress the impact of the physical factors.

		\subsection{Evaluation in Virtual World}
		\label{Attack on Virtual-World}
		Following the setup in the digital world, we select different human models as the attacked object. Without loss of generality, we choose background as the poisoned label to fool detectors in our experiment. We use four different physical factors illustrated in Fig.~\ref{simulateworld} for evaluating the efficacy of the proposed RBK.
		\begin{table*}[!t]
			\centering
			\tabcolsep=2.5pt
			\renewcommand{\arraystretch}{0.7}
			\setlength{\abovecaptionskip}{0pt}
			\setlength{\belowcaptionskip}{0pt}
			\caption{\label{table_simulateworld} The attack performance of detectors in different virtual scenes.}
			\begin{tabular}{ccccccccccccccccc}
				\toprule[2pt]
				\multirow{2}{*}{Object detector} & &  \multicolumn{3}{c}{Distance}     & &\multicolumn{3}{c}{Rotation}  & & \multicolumn{3}{c}{Brightness} & & \multicolumn{3}{c}{Brightness} \\ \cmidrule(r){3-5}\cmidrule(r){7-9}\cmidrule(r){11-13}\cmidrule(r){15-17}
				& & 20 & 70 & 120 &  & 0 & 30 & 60 & & 0.1 & 10 & 100 &  & 50 & 100 & 150                    \\\cmidrule(r){1-17}
				YOLOv5s & & 0.00 & 0.00 & 8.33 & & 0.00 & 0.00 & 0.00 & & 4.16& 0.00 & 0.00 & & 0.00 & 0.00 & 0.00 \\
				Baseline & & 91.66 & 16.66 & 12.50 & & 91.66 & 20.83 & 4.16 & & 4.16 & 79.16 & 8.33 & & 100.00 & 75.00 & 33.33    \\
				$\mathit{RD}$ (Ours)& & \textbf{100.0} & \textbf{83.33} & \textbf{79.16} &  & \textbf{100.0} & \textbf{75.00} & \textbf{29.16} & & \textbf{87.5} & \textbf{95.83} & \textbf{58.33} & & \textbf{100.00} & \textbf{83.33} & \textbf{66.66}   \\
				\bottomrule[2pt]
			\end{tabular}
		\end{table*}
		
		\begin{figure*}[!t]
			\centering
			\setlength{\abovecaptionskip}{0pt}
			\setlength{\belowcaptionskip}{0pt}
			\includegraphics[scale=0.24]{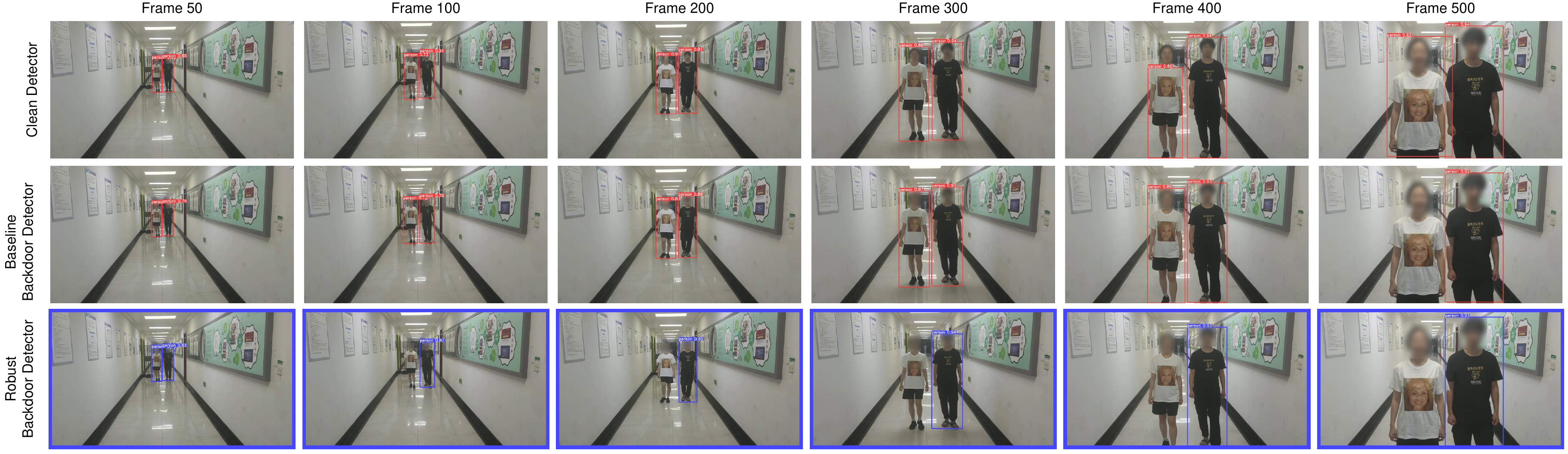}
			\caption{\label{fig_real_1}Visualization of different detectors indoors. The figure shows the effectiveness of trigger T-shirt for person to evading the backdoor object detector indoors. Each row corresponds to various detectors while each column shows an individual frame in a video.}
		\end{figure*}
		\begin{figure*}[!t]
			\centering
			\setlength{\abovecaptionskip}{0pt}
			\setlength{\belowcaptionskip}{0pt}
			\includegraphics[scale=0.21]{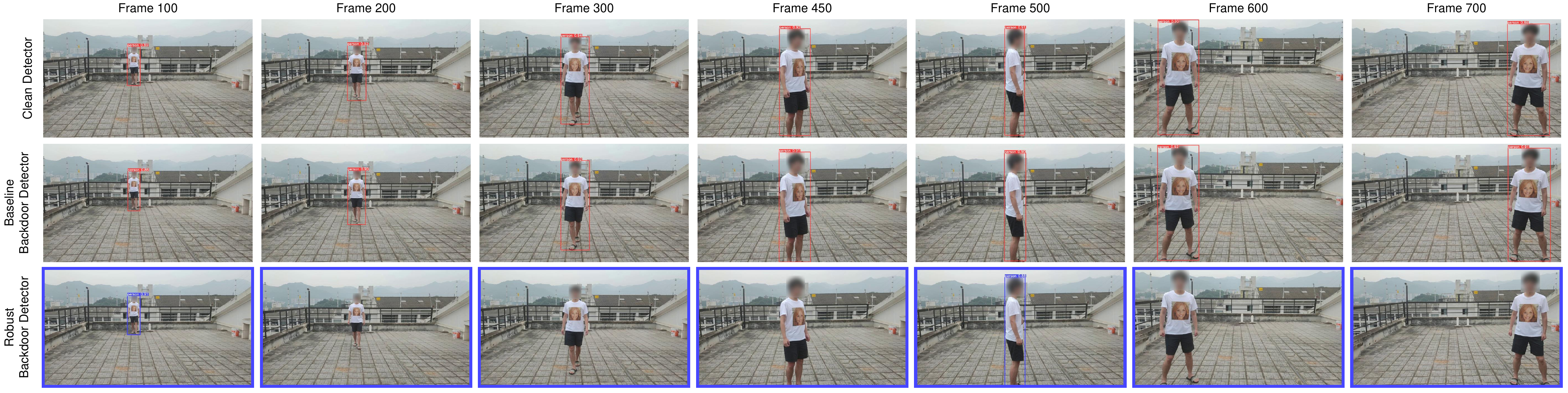}
			\caption{\label{fig_real_2}Visualization of different detectors outdoors in sunny.The figure shows the effectiveness of trigger T-shirt for person to evading the backdoor object detector outdoors.}
		\end{figure*}
		\begin{figure*}[!t]
			\centering
			\setlength{\abovecaptionskip}{0pt}
			\setlength{\belowcaptionskip}{0pt}
			\includegraphics[scale=0.21]{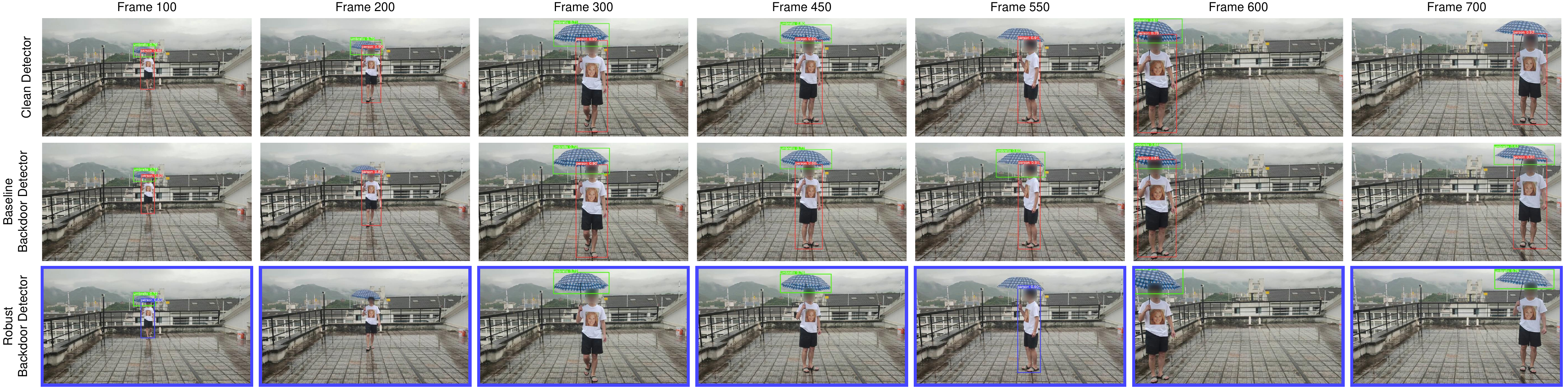}
			\caption{\label{fig_real_3}Visualization of different detectors outdoors in rain. The figure shows the effectiveness of trigger T-shirt for person to evading the backdoor object detector in rain.}
		\end{figure*}

		As shown in Fig.~\ref{fig_sim_dis}, we find that the higher rotation angle and distance can make our $\mathit{RD}$ less effective. This can be attributed to the fact that the trigger is badly captured when the pixel information of the trigger is lost. We adjust the light sources to simulate the different daily times indoors and outdoors in Fig.~\ref{fig_sim_light}. As the light condition is brighter or darker, the person with the trigger also is undetected, which indicates that the simple light change is invalid for our $\mathit{RD}$. Fig.~\ref{fig_sim_rain} shows that $\mathit{RD}$ is robust to the rain. But when the number of raindrops comes to 150 and 200, the person with the trigger is partially occluded in the sight of the detector. Tab.~\ref{table_simulateworld} shows that the percentage of misdetection in the virutal world. We find that our attack method outperforms the previous work on backdoor object detector. In addition, the robustness of $\mathit{RD}$ is poor in the brighter scenes. This can be attributed to the fact that too bright envirnoment can damage the most feature of the triggeer captured badly by backdoor object detector. 
		
		We find the performance of $\mathit{RD}$ is better than normal backdoor object detection when exposed to most physical factors in the virtual world.
		
		\begin{table*}[!t]
			\centering
			\tabcolsep=2.5pt
			\renewcommand{\arraystretch}{1.0}
			\setlength{\abovecaptionskip}{0pt}
			\setlength{\belowcaptionskip}{0pt}
			\caption{\label{table_realworld} The attack performance of detectors in different real scene.}
			\begin{tabular}{ccccccccccccc}
				\toprule[2pt]
				\multirow{2}{*}{Object detector} & &  \multicolumn{3}{c}{Indoor, Sunny}     & &\multicolumn{3}{c}{Outdoor, Sunny}  & & \multicolumn{3}{c}{Outdoor, Rainy} \\ \cmidrule(r){3-5}\cmidrule(r){7-9}\cmidrule(r){11-13}
				& & Video 1 & Video 2 & Video 3 &  & Video 4 & Video 5 & Video 6 & & Video 7 & Video 8 & Video 9                     \\\cmidrule(r){1-13}
				YOLOv5s & & 5.02 & 1.35 & 2.68 & & 3.70 & 2.56 & 0.59 & & 2.58 & 0.63 & 1.27\\
				Baseline & & 17.63 & 21.84 & 6.96 & & 5.25 & 3.47	 & 4.86 & & 5.02 & 4.41 & 6.11     \\
				$\mathit{RD}$ (Ours)& & \textbf{67.65} & \textbf{69.01} & \textbf{40.87} &  & \textbf{62.56} & \textbf{61.42} & \textbf{35.94} & & \textbf{61.42} & \textbf{78.10} & \textbf{66.88}    \\
				\bottomrule[2pt]
			\end{tabular}
		\end{table*}
		\subsection{Evaluation in Real World}
		\label{Attack on Real World}
		To evaluate the effectiveness of $\mathit{RD}$ in the real physical world, we make a T-shirt with a $Face$ trigger. During the physical experiment in the real world, we use a smartphone to record several pieces of video. Because the distance between the viewing point and attacked object can influence the success rate of a backdoor attack, we examine the impact of varying distances and angles in Figures~\ref{fig_real_1}, ~\ref{fig_real_2} and ~\ref{fig_real_3}. The Tab.~\ref{table_realworld} also shows the percentage of the misdetection in video set. It shows that the smaller the distance, the higher the success rate. Generally, a higher success rate is achieved at narrow angles than wide angles, and at short distances than long distances. Since $\mathit{RD}$ has the stronger capability to recognize triggers as the distance comes smaller, it is easier for $\mathit{RD}$ to hide the $B$-box. The performance of the baseline backdoor object detector is weaker than $\mathit{RD}$ when facing numerous physical factors. Once the proposed $RBK$ method has been exploited during the training of object detector, the performance of backdoor object detector are greatly improved.
		
		\begin{table*}[!t]
			\centering
			\renewcommand{\arraystretch}{1.0}
			\setlength{\abovecaptionskip}{0pt}
			\setlength{\belowcaptionskip}{0pt}
			\caption{\label{size} The performance (\%) of the trained backdoor object detector by different trigger sizes on the three COCO datasets. The trigger transparency is fixed to 1.}
			\begin{tabular}{cccccccc}
				\toprule[2pt]
				Method & Trigger Size & $AP_b$ & $mAP_b$  & $mAP_a$ & $AP_{a+b}$ & $mAP_{a+b}$ & $ASR$       \\ \hline
				\multirow{6}{*}{\begin{tabular}[c]{@{}c@{}} Invariable\\ Trigger size\end{tabular}} 
				& $20\times 20$ & \textbf{75.4} & 55.2 & 51.8 & 50.0 & 51.8 & 50.42                                                          \\
				& $40\times 40$    & 74.1 & 55.1 & 51.9 & 52.7 & 51.9 & 39.17                                                       \\
				& $60\times 60$    & 74.7 & 55.2 & 52.0 & 53.5 & 52.0 & 40.49                                                         \\
				& $80\times 80$    & 75.2 & \textbf{55.3 }& 52.2 & 55.6 & 52.2 & 39.48                                                           \\
				& $100\times 100$   & 75.1 & \textbf{55.3} & 52.2 & 57.7 & 52.2 & 33.34                                                           \\
				& $120\times 120$    & 75.1 & 55.1 & 52.1 & 59.8 & 52.2 & 31.15                                                            \\
				\hline
				\begin{tabular}[c]{@{}c@{}} Variable\\ Trigger size\end{tabular} & -- & 74.1 & 54.9 & \textbf{52.5} & \textbf{22.6} & \textbf{52.2} & \textbf{85.86} \\
				\bottomrule[2pt]
			\end{tabular}
		\end{table*}
		
		\begin{figure*}[!t]
			\centering
			\setlength{\abovecaptionskip}{0pt}
			\setlength{\belowcaptionskip}{0pt}
			\includegraphics[scale=0.43]{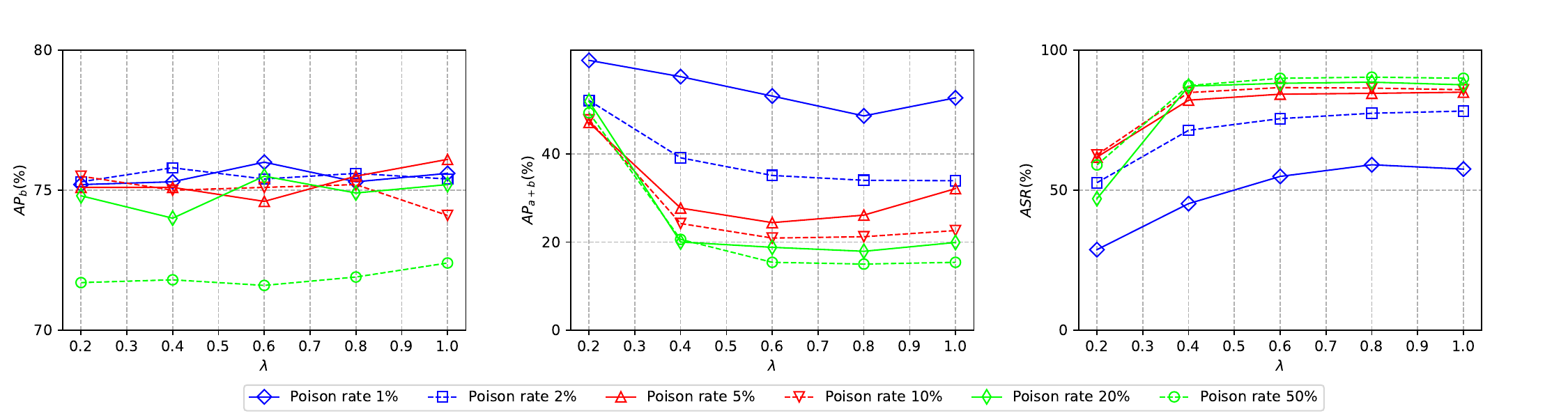}
			\includegraphics[scale=0.43]{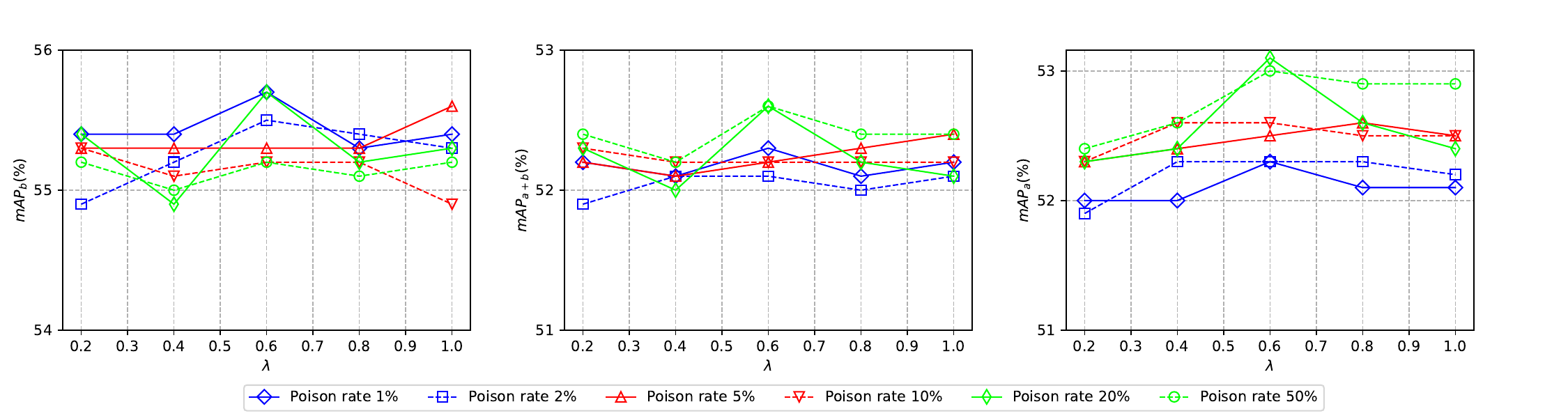}
			\caption{\label{fig5}The clean accuracy and $ASR$ of the backdoor object detector with different poison rate $Poi$ and transparency $\lambda$.}
		\end{figure*}
		
		\subsection{Ablation Analysis}
		\label{Ablation Analysis}
		In this section, we do a series of experiments on the COCO dataset. We study the impact of different parameters on $ASR$ and the clean accuracy of the backdoor object detector: (1) backdoor trigger size, (2) transparency $\lambda$, (3) backbone, and (4) loss function $L_v$. In the next experiments, we set the trigger as $Face$. Also, for the study on each parameter of RBK, we fix the other parameters.

		\textbf{Ablation Study on Trigger Size.} 
		Here, we show how the invariable-size triggers impact the clean accuracy and $ASR$ of the backdoor object detector. Different trigger sizes consist of $20\times20$ to $120\times120$ at 20 intervals. As shown in Tab.~\ref{size}, the $20\times20$ size of the trigger achieve 75.4\% $AP_b$ of the “person” class and 50.42\% $ASR$. With the size of the trigger from $20\times20$ to $120\times120$, $ASR$ of the backdoor attack decreases from 50.42\% down to 31.15\%. In contrast, the variable-size trigger reaches 85.86\% $ASR$. Despite the changes of $AP_a$ and $mAP_b$, the clean accuracy of the variable-size trigger is not much different from the others. This ablation study shows that the performance improvement using variable-size triggers is significant.
		
		\begin{table*}[!t]
			\centering
			\renewcommand{\arraystretch}{1.0}
			\setlength{\abovecaptionskip}{0pt}
			\setlength{\belowcaptionskip}{0pt}
			\caption{\label{backbone} Training YOLOv5s with different backbone, clean accuracy (\%) and $ASR$ (\%) of backdoor attacks with 10\% poisoning rate and transparency of 1.}
			\begin{tabular}{cccrrrrcr}
				\toprule[2pt]
				Object detector & Backbone    & Poison Rate     & $AP_b$      & $mAP_b$  & $mAP_a$ & $AP_{a+b}$ & $mAP_{a+b}$     & $ASR$       \\ \hline
				\multirow{10}{*}{YOLOv5s}&\multirow{2}{*}{DarkNet} & 10\% & 74.1 & 54.9 & 52.5 & 22.6 & 52.2 & 85.86                                                          \\
				& & 0\%    & 76.4 & 56.7 & 52.7 & 60.5 & 52.8 &
				--                                                       \\
				\cmidrule(r){2-9}
				&\multirow{2}{*}{VGG11~\cite{[64]}}& 10\%    & 76.3 & 55.5 & 52.5 & 23.0 & 52.5 & 84.82                                                          \\
				& & 0\%     & 76.2 & 55.5 & 51.8 & 60.1 & 51.9 & --                                                          \\
				\cmidrule(r){2-9}
				&\multirow{2}{*}{ResNet50~\cite{[2]}}& 10\%     & 76.9 & 59.4 & 57.2 & 18.5 & 56.7& 87.15                                                           \\
				& & 0\%     & 76.6 & 58.7 & 55.2 & 61.3 & 55.2 & --                                                            \\
				\cmidrule(r){2-9}
				&\multirow{2}{*}{MobileNetV3~\cite{[65]}}& 10\%     & 68.0 & 47.1 & 44.9 & 22.4 & 44.7 & 84.17                                              \\
				& & 0\%     & 68.7 & 46.8 & 44.0 & 56.8 & 44.2 & --                                                      \\
				\cmidrule(r){2-9}
				&\multirow{2}{*}{DenseNet121~\cite{[66]}}& 10\%     & 75.7 & 57.0 & 54.7 & 22.2 & 54.3 & 86.68                                                         \\
				& & 0\%     & 76.2 & 56.5 & 52.9 & 54.3 & 52.6 & --                                                     \\
				\bottomrule[2pt]
			\end{tabular}
		\end{table*}
		
		\textbf{Ablation Study on $\lambda$.} Whether the triggers are human-perceived during training is a critical issue. Transparency enhances the invisibility of the triggers when added to the training dataset. Fig.~\ref{fig5} shows the variation of clean accuracy and $ASR$ on different transparency $\lambda$, and poison rate $Poi$. In general, the larger $\lambda$ is, the easier for a backdoor to attack the detector, and the bigger $ASR$ of the backdoor attack is. Of interest are the results for the backdoor object detector, $ASR$ decrease a little from 85\% to 81\% during $Poi>=5\%$ and decrease rapidly to 53\% with only a slight drop on $Poi$. It can be seen that the change of both parameters will have little influence on the clean accuracy except when the poison rate is 50\%, the level of poison rate makes serious damage to $AP_b$ which reduces by 3\%. Fig.~\ref{fig5} also shows $mAP_b$ and $mAP_{a}$ illustrating that poison rates and transparency have little effect on the average accuracy of other classes, which fluctuate between 0.4\%. The sensitivity of its $mAP_{a+b}$ to the poison rate and transparency is even weaker in the dataset $D_{val,a}$. Given the above, $Poi$ and $\lambda$ also make an impact on $ASR$ and have little influence on clean accuracy.
		
		
		\textbf{Ablation Study on Backbone.} The structure of an object detector is usually composed of different components. The extracted feature of the trigger depends on the backbone which is one component of the object detector. So we show whether the different backbones affect the performance of the backdoor attack. We compare different backbone VGG11~\cite{[64]}, Resnet~\cite{[2]}, MobileNetV3~\cite{[65]}, DenseNet121~\cite{[66]} for ablation experiments in Tab.~\ref{backbone}. 
		
		The clean accuracy and $ASR$ of five backbones are ordered from smallest to largest as MobileNetV3, VGG11, DarkNet, DenseNet121, and ResNet50. Darknet is the backbone that has small parameter quantities achieving large accuracy. It can be seen that the larger parameter quantities which strengthen the model's ability to recognize the trigger will cause the model easier to be attacked more. ResNet50 which makes the trigger better recognized outperforms another backbone with 2.5\% higher clean accuracy and 1.3\% higher $ASR$ than Darknet. But the training time increases to 2.15 times and the inference time increases to 1.38 times at the cost of increased accuracy.
		
		\begin{table}[!t]
			\centering
			\renewcommand{\arraystretch}{1.0}
			\setlength{\abovecaptionskip}{0pt}
			\setlength{\belowcaptionskip}{0pt}
			\caption{\label{table3}Clean accuracy (\%) and $ASR$ (\%) of Robust Backdoor training using different loss function $L_y$ and $L_v$.}
			\setlength{\tabcolsep}{0.1cm}{
				\begin{tabular}{cccccccc}
					\toprule[2pt]
					Method    & $AP_b$      & $mAP_b$ & $mAP_a$  & $AP_{a+b}$ & $mAP_{a+b}$     & $ASR$       \\ \hline
					YOLO\ $L_v,L_y$  & 74.1 & 54.9 & 52.5& 22.6 & 52.2 & 85.86                                                          \\
					\hline
					YOLO\ w/o\ $L_v$   & 75.8 & 55.8 & 52.7  & 31.9 & 52.7 & 80.12                                                           \\
					\hline
					YOLO\ w/o\ $L_y$    & 73.7 & 54.7 & 52.4 & 20.1 & 51.9 & 87.15                                                           \\
					\bottomrule[2pt]
			\end{tabular}}
		\end{table}
		
		\textbf{Ablation Study on $L_v$ and $L_y$.} The strong perturbations may cause interference on the clean detector and affect the backdoor attack, so we introduce $L_v$ and $L_y$ to improve the clean accuracy and $ASR$. Tab.~\ref{table3} shows that the performance of the backdoor object detector without $L_v$ or $L_y$ decreases on both clean and poisoned images. Without $L_v$, $ASR$ of robust backdoor object detector $\mathit{RD}$ decrease by 5.75\%, and the $AP_b$ increase by 1.7\% respectively. The loss function $L_v$ makes the stronger perturbation to be learned by the detector so that $\mathit{RD}$ has a greater $ASR$. Without $L_y$, the clean accuracy of the backdoor object detector decrease by 0.4\%, and $ASR$ increase by 1.29\%.

		
		\begin{table*}[!t]
			\centering
			\tabcolsep=2.5pt
			\renewcommand{\arraystretch}{0.7}
			\setlength{\abovecaptionskip}{0pt}
			\setlength{\belowcaptionskip}{0pt}
			\caption{\label{bbox number} The percentage of $B$-box (\%) detected as the person class by clean detector, Backdoor object detector, and Robust backdoor object detector with different datasets. Low represents $Score_B \in [0,0.1]$. Middle represents $Score_B \in (0.1,0.5]$. High represents $Score_B \in (0.5,1.0]$. }
			\begin{tabular}{ccccccccccccccc}
				\toprule[2pt]
				\multirow{2}{*}{Object detector} & & Dataset & &  \multicolumn{3}{c}{Clean}     & &\multicolumn{3}{c}{Backdoor}  &    & \multicolumn{3}{c}{Backdoor with $\Delta_{phy}$}     \\ \cmidrule(r){5-7}\cmidrule(r){9-11}\cmidrule(r){13-15}
				& & $Score_B$ & & Low & Middle & High &  & Low & Middle & High & & Low & Middle & High                     \\\cmidrule(r){2-15}
				YOLOv5s & & \multirow{3}{*}{\begin{tabular}[c]{@{}c@{}} $B$-box \\ Quantity \end{tabular}}& & 99.55 & 0.31 & 0.14 & &
				99.43 & 0.46 & 0.11 & & 99.49 & 0.41 & 0.10\\
				Baseline & &  & & 99.35 & 0.49 & 0.16 & & 99.68 & 0.30 & 0.02 & & 99.53 & 0.38 & 0.09     \\
				$\mathit{RD}$ (Ours)& & & & 99.36 & 0.48 & 0.16 &  & 99.68 & 0.29 & 0.03 & & 99.64 & 0.32 & 0.04    \\
				\bottomrule[2pt]
			\end{tabular}
		\end{table*}
		\section{Discussion}
		\label{Discussion of RBA}
		In this section, we discuss the possible reasons why our attack method is effective. When disturbed by the physical factor occurring in the real world, the trigger will lose connection with backdoor-related neurons of backdoor object detection. The blocked feature of the trigger will be damaged which can respond to the backdoor-related neurons. Therefore, the original $B$-box with the higher $Score_B$ will be retained by NMS. We can observe that the detection of the object with the trigger is $y$, not $\hat{y}$.
		
		Therefore, the adversary hopes the backdoor object detector learns about more different sizes of triggers with more physical noises. By learning the feature of this trigger hard to be extracted by the backbone, $\mathit{RD}$ reduces the sensitivity to physical factors. The backdoor-related neurons will be activated which affect the generation of the original $B$-box by lowering the $Score_B$ of the original $B$-box seen by Tab.~\ref{bbox number}. Therefore, the number of $B$-box closed to $y$ will be limited and the number of $B$-box closed to $\hat{y}$ will be retained by NMS. As a result, the object with the trigger disturbed by physical factors will be detected as $\hat{y}$. 
		
		To summarize, our RBK increases the diversity of backdoor triggers, making the backdoor object detector strengthen the association between these triggers and poisoned labels. It widens the boundary where an object with a trigger will be detected as a poisoned label. Even if the physical factors are added to the trigger, the object with the trigger will be near the boundary but still stay in the boundary of $\mathit{RD}$.

		\section{Conclusion}
		This paper introduces the robust backdoor attack on object detectors. We report the disadvantages of existing backdoor attacks on object detection, which lack robustness to physical factors and improve the performance only in the digital world. We first propose a variable-size trigger that can fit different sizes of attacked objects to reflect the distance between the viewing point and attacked object in the real world. In addition, to enhance the robustness of the backdoor object detector on physical factors, we propose malicious adversarial training to adapt the detector to most physical factors, for which we generate the strongest physical noise example hard to be detected by the detector in the pixel space. Experiments demonstrate that our method performs effectively in the digital, virtual, and real worlds. We also demonstrate that our method ensures the robustness of the backdoor object detector on physical noises and vertical rotation of objects. In summary, security-sensitive domains need to focus on the security threat of robust backdoor attacks. Backdoor attacks on object detection robust to the real world can be harmful to humans.
		
		\section*{Acknowledgment}
		This work was supported by the National Natural Science Foundation of China (92167203), Zhejiang Provincial Natural Science Foundation of China (LZ22F020007), National Natural Science Foundation of China (62202457), China Postdoctoral Science Foundation (2022M713253), and Science and Technology Innovation Foundation for Graduate Students of Zhejiang University of Science and Technology (F464108M05).
		
		\bibliography{mybibfile}

	\end{document}